\documentclass[letterpaper, 10 pt, conference]{ieeeconf}  

\usepackage[table]{xcolor}

\usepackage{multirow}
\usepackage{times}

\usepackage{hyperref}

\usepackage{color}
\usepackage{algorithm}
\usepackage{algpseudocode}
\usepackage{listings}
\usepackage{courier}
\usepackage{soul}
\usepackage{amsmath,amsfonts,amssymb}

\usepackage{mathtools}
\usepackage{subfigure}
\usepackage{wrapfig}
\usepackage{hyperref}
\usepackage{multirow}
\usepackage{graphicx}
\usepackage{graphicx}
\usepackage{soul}
\usepackage[flushleft]{threeparttable}
\usepackage{enumerate}
\usepackage{algorithmicx}
\usepackage{algpseudocode}
\usepackage{epstopdf}

\IEEEoverridecommandlockouts                              
\definecolor{red}{rgb}{1.00,0.00,0.00}
\definecolor{blue}{rgb}{0.00,0.00,1.00}
\definecolor{green}{rgb}{0.2,0.70,0.2}
\definecolor{yellow}{rgb}{0.5,0.5,0.0}
\definecolor{white}{rgb}{1,1,1}

\title{\LARGE \bf
A Robust Biped Locomotion Based on Linear-Quadratic-Gaussian Controller and Divergent Component of Motion
\author{Mohammadreza Kasaei, Nuno Lau and Artur Pereira
 \\IEETA / DETI University of Aveiro 3810-193 Aveiro, Portugal \\
	\{mohammadreza, nunolau, artur\}@ua.pt
}
}

\begin{document}
\maketitle
\thispagestyle{empty}
\pagestyle{empty}
\begin{abstract}	
Generating robust locomotion for a humanoid robot in the presence of disturbances is difficult because of its high number of degrees of freedom and its unstable nature. In this paper, we used the concept of Divergent Component of Motion~(DCM) and propose an optimal closed-loop controller based on Linear-Quadratic-Gaussian to generate a robust and stable walking for humanoid robots. The biped robot dynamics has been approximated using the Linear Inverted Pendulum Model~(LIPM). Moreover, we propose a controller to adjust the landing location of the swing leg to increase the withstanding level of the robot against a severe external push. The performance and also the robustness of the proposed controller is analyzed and verified by performing a set of simulations using~\mbox{MATLAB}. The simulation results showed that the proposed controller is capable of providing a robust walking even in the presence of disturbances and in challenging situations.
\end{abstract}
\let\thefootnote\relax\footnotetext{This paper has been accepted for publication in the Proceedings of the 2019 IEEE/RSJ International Conference on Intelligent Robots and Systems (IROS 2019).} 

\textbf{Keywords:}
Biped locomotion, Humanoid robot, Robust walk engine, Linear inverted pendulum, Linear-Quadratic-Gaussian~(LQG).
\vspace{-4mm}
\section{Introduction}
\label{sec:introduction}
Walking performance of humanoid robots has been much improved but it is still far from the expectations. Although many humanoid robots are able to perform walking on flat terrain (or on known geometry uneven terrain), just a few of them can compensate severe disturbances or walk on rough terrain. A humanoid robot has a distinctive property which is the similarity in kinematics as well as dynamics to a human. According to this property, they can adapt to our daily life environment without changing it and humans expect them to be able to step out of the laboratory and perform their daily life tasks and also precisely and safely collaborate with them. The critical requirement for a humanoid robot to work in our dynamic environment is the ability to react robustly against unknown disturbances while walking. Nowadays the number of researches in the field of developing a robust walk engine is increasing and can be categorized into two main categories: model-based and model-free. In model-based approaches, a physical model for the dynamics of the system is considered and based on this model, a walking system is designed. On the other hand, model-free approaches try to design a walking system just by generating a rhythmic motion for each limb without any dynamics model. The main focus of this paper is on a model-based approach.
\begin{figure}[!t]
	\label {Hirachical}
	\centering
	\includegraphics[width =1\linewidth, trim= 0.5cm 0.3cm 0cm 0cm,clip]{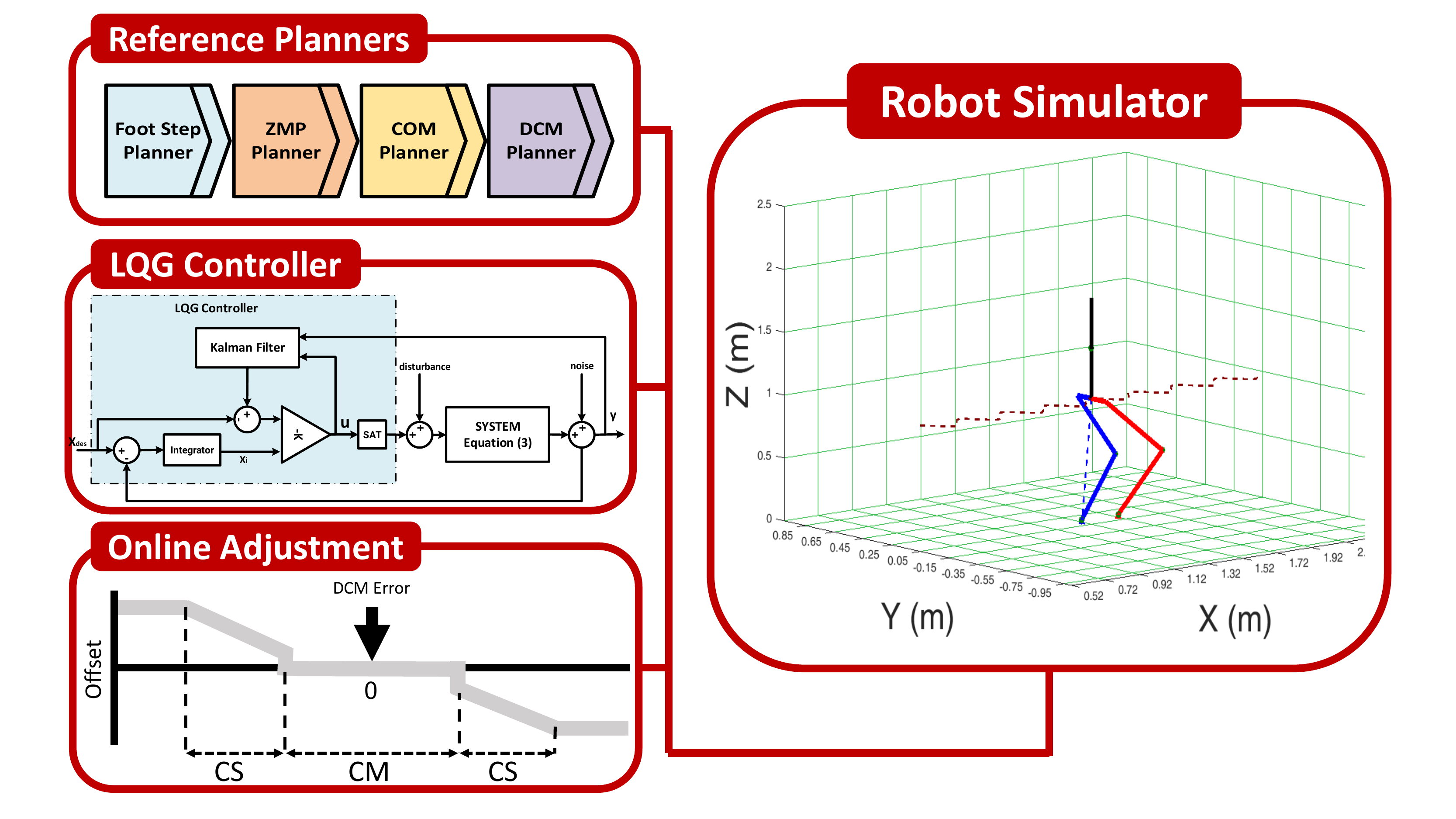}
	\vspace{-5mm}
	\caption{ An overview of the proposed walking system. The architecture of the system is composed of four main modules which are: reference planners, LQG controller, online adjustments and robot simulator.}
	\vspace{-5mm}
	\label{fig:Hirachical}
\end{figure}
Although considering the full rigid-body dynamics of the robot is not impossible, it is typically computationally expensive and its solving time is too long due to dealing with the high nonlinear as well as big dimensional problems. Hence, considering a full body dynamics model is not affordable for real-time implementation. Instead, for reducing the computation cost and complexity of the planning process, the overall dynamics is restricted to the center of mass~(COM). Linear Inverted Pendulum Model~(LIPM) is one of the well-known models which has been proposed by Kajita and Tani~\cite{kajita1991study}. This model has been commonly used to approximate the overall dynamics of a humanoid. The popularity of this model comes from its fast and efficient solution for real-time implementation. This model approximates the dynamics of the robot just by considering a single mass which is restricted to move along a horizontally defined plane and connected to the ground via a massless rod. Based on these assumptions, a straightforward solution exist to generate the reference COM trajectories according to a set of planed footsteps. The planned trajectories guarantee long-term stability and a controller can be used to produce walking by tracking these references via feedback control.

In this paper, we formulate the problem of generating a dynamic walk for humanoids as a closed-loop optimal control algorithm and by performing some simulations, we show how this controller could generate a robust walking for a humanoid even in very challenging situations. Furthermore, we propose a controller to adjust the landing location of the swing leg to increase the withstanding level of the robot. This paper is structured as follows: Section~\ref{sec:related_works} gives an overview of related work. In Section~\ref{sec:DynamicsModel}, the fundamentals of the presented approach is explained and show how the reference trajectories are generated. The overall architecture of the proposed controller and also its robustness are discussed in Section~\ref{sec:Control}. An online foot step adjustment method is presented in Section~\ref{sec:online_footstep} and show how it could improve the push recovery capability of the controller. Finally, conclusions and future research are presented in Section~\ref{sec:CONCLUSION}.
\vspace{-2mm}
\section {Related Work}
\label{sec:related_works}
Recent advances in humanoid robot locomotion have been based on decoupling COM dynamics into divergent and convergent components. Several kinds of research ~\cite{takenaka2009real,englsberger2015three,englsberger2017smooth,kamioka2018simultaneous} showed that a robust stable walking could be developed just by controlling the divergent component of motion. In the remainder of this section we briefly review some of these approaches.

Pratt et al.~\cite{pratt2006capture} extended LIPM model replacing the single mass with a flywheel (also called a reaction wheel) to consider the angular momentum around COM. Based on this model, they define the Capture Point~(CP) which is conceptually a point on the ground that the robot should step to keep its stability. Stephens~\cite{stephens2007humanoid} used their model to determine decision surfaces that could determine when a particular recovery strategy (e.g., ankle, hip or step) should be used to regain balance. Later, Morisawa et al.~\cite{morisawa2014biped} utilized information of the support region to generate online the walking patterns and proposed a PID balance controller based on CP to keep the stability of a robot while walking on uneven terrain. They showed the effectiveness of their controller using several simulations with the HRP-2 humanoid robot.

Kryczka et al.~\cite{kryczka2015online} introduced an algorithm for online planning of bipedal locomotion which was based on a nonlinear optimization technique. Their method tries to find a set of step parameters including step position and step time which could control the robot from the current state to the desired state. They applied their method on the real humanoid platform COMAN and performed a set of experiments to show how the step time modification could extend the robustness of the controller and improve the capability to recover stability regardless of the disturbance source.

Takaneka et al.~\cite{takenaka2009real} were the first ones that proposed DCM concept and used it to plan and real-time control of a humanoid walking. They showed that their method is able to generate all the walking trajectories every 5$ms$ which was enough to react robustly against unknown obstacles and unwanted disturbances. Besides, according to their simulation results, they showed that to develop stable locomotion just the divergent component needs to be controlled.

Englsberger et al.~\cite{englsberger2015three} extended DCM to 3D and introduced Enhanced Centroidal Moment Pivot point~(eCMP) and also the Virtual Repellent Point~(VRP) which could be used to encode the direction, magnitude and total forces of the external push. Moreover, they showed how eCMP, VRP, and DCM can be used to generate a robust three-dimensional bipedal walking. The capabilities of their approaches have been verified both in simulations and experiments. 

Khadiv et al.~\cite{khadiv2016stepping} proposed a method based on combining DCM tracking and step adjustment to stabilize biped locomotion. They used LIPM and a set of planned footstep to generate DCM reference trajectories and also they used a Hierarchical Inverse Dynamics~(HID) to apply the generated trajectories to a real robot. Moreover, they proposed a strategy to adjust the swing foot landing location based on DCM measurement. They carried out some simulations using active and passive ankles simulated robots to demonstrate the effectiveness of their method.

Hopkins et al.~\cite{hopkins2014humanoid} introduced a novel framework for dynamic humanoid locomotion on uneven terrain using a time-varying extension to the DCM. They argued that by modifying the natural frequency of the DCM, the generic COM height trajectories could be achieved during stepping. They showed that their framework could generate DCM reference trajectories for dynamic walking on uneven terrain given vertical COM and Zero Momentum Point~(ZMP) trajectories~\cite{vukobratovic1970stability}. Their method has been verified in simulation using a simulated humanoid robot.

Griffin et al.~\cite{griffin2016model} used the time-varying DCM concept and formulated the problem of generating a stable walking as a Model Predictive Control~(MPC). In their method, the step positions and rotations were used as control inputs which allowed to generate a fast and robust walking for a humanoid robot. They showed the performance of their controller using some simulations with the ESCHER humanoid and the results demonstrated the effectiveness of their approach. 

Shafiee-Ashtiani et al.~\cite{shafiee2017robust} also formulated robust walking as a MPC which was based on time-varying DCM concept. They showed the performance of the proposed walking using some simulation scenarios. The simulation results showed that their controller was able to keep the stability of the simulated robot in various situations.

Kamioka et al.~\cite{kamioka2018simultaneous} analytically derived a solution of DCM for a given arbitrary input function and according to this solution, they proposed a novel quadratic programming~(QP) to optimize ZMP and also foot placements simultaneously. They demonstrated the performance of the proposed algorithm by conducting a push recovery experiment using a real humanoid robot. The results showed that their method could compensate disturbances in a hierarchical strategy.

Most of the above-mentioned works use online optimization methods like QP or MPC to plan and control the patterns of walking. Although they have been quite successful, in comparison with feedback control methods, they are not versatile~\cite{griffin2016model,carpentier2016versatile}. These types of methods are commonly based on iterative algorithms, hence their performance is sensitive to the resources computation power. Therefore, in such methods, the Reaction Times~(RT) are totally dependent on computation power. Since RT is a critical parameter to react robustly against severe push and reducing the time taken to detect the push and re-plan the trajectories increases the probability of regaining stability. 
In the rest of this paper, to decrease the reaction time, we decouple the control and re-planning procedures. Particularly, we formulate the tracking procedure of walking as a Linear-Quadratic-Gaussian~(LQG) controller which is a robust offline optimal state-feedback controller and we combine a step adjustment controller to LQG controller for adjusting the upcoming steps to improve the robustness.

\section{Dynamics Model and Reference Planning}
\label{sec:DynamicsModel}
In this section, we briefly review the LIPM and DCM dynamics which are the fundamentals of our approach. We explain how they can be used to represent the overall dynamics of a humanoid in a state space form. Moreover, we illustrate the procedure of reference trajectories planning. It should be noted that since all the equations in sagittal and frontal planes are equivalent and independent, we derive the equations just in the sagittal plane.

\subsection{Linear Inverted Pendulum Model~(LIPM)}
This model approximates the robot dynamics by considering a restricted dynamics of the COM. According to this model, COM is limited to move along a predefined horizontal plane, therefore, the motion in sagittal and frontal planes are decoupled and also independent. This model represents the overall dynamics of a humanoid robot by a first-order stable dynamics as follow:
\begin{equation}
\ddot{x} = \omega^2 ( x - p_x) \quad ,
\label{eq:lipm}
\end{equation}
\noindent
where $x$ represents the position of COM in the sagittal plane, $\omega = \sqrt{\frac{g+\ddot{z}}{z}}$ is the natural frequency of the pendulum, $p_x$ is the position of ZMP which is a point on the ground plane where the ground reaction force acts to compensate gravity and inertia~\cite{vukobratovic1970stability}. 
\subsection{Divergent Component of Motion~(DCM)}
The unstable part of COM dynamics is called DCM and conceptually it is the point that robot should step to come to rest over the support foot~\cite{pratt2006capture}. The DCM dynamics is defined as follow:
\begin{equation}
\zeta = x + \frac{\dot{x}}{\omega} \quad ,
\label{eq:dcm}
\end{equation}
\noindent
where $\zeta$, $\dot{x}$ represent the DCM and the velocity of COM respectively. By taking the derivative from both sides of this equation ($\dot{\zeta} = \dot{x} + \frac{\ddot{x}}{\omega}$) and substituting Equation~\ref{eq:lipm} into this equation, the dynamics of the LIPM can be represented using a state space system as follows:
\begin{equation}
\frac{d}{dt} \begin{bmatrix} x \\ \zeta \end{bmatrix}
= 
\begin{bmatrix} 
-\omega & \omega \\ 
0 & \omega \\ 
\end{bmatrix}	
\begin{bmatrix} x \\ \zeta \end{bmatrix}
+
\begin{bmatrix} 
0 \\
-\omega
\end{bmatrix} p_x \quad .
\label{eq:statespace_zeta}
\end{equation}

According to this system, COM is always converging to the DCM without any control. Therefore a stable walking can be developed just by controlling the DCM. 

\subsection{Reference trajectories planning}
\begin{figure}[!t]
	\label {Planner}
	\centering
	\includegraphics[width = 0.85\columnwidth, trim= 0.1cm 7cm 1cm 7cm,clip]{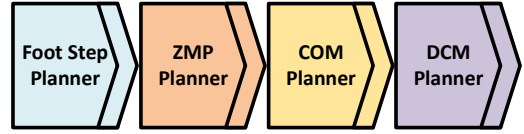}
	\caption{ Procedure flow of reference trajectories generators which is composed of four independent planners.}
	\label{fig:Planner}
\end{figure}

The overall architecture of our reference trajectories planner is depicted in Fig.~\ref{fig:Planner}. As is shown in this figure, it is composed by four planners which are \textit{Foot Step Planner}, \textit{ZMP Planner}, \textit{COM Planner} and \textit{DCM Planner}. Foot Step Planner generates a set of footstep positions according to the input step info which are Step Length~(SL), Step Width~(SW), Step Duration~(SD) and Single Support Duration~(SSD) and Double Support Duration~(DSD). ZMP planner generates ZMP trajectories based on the specified footsteps and using the following formulation:
\begin{equation}
p_x= 
\begin{cases}
f_{i,x}  \qquad\qquad\qquad\qquad\qquad 0 \leq t < T_{ss} \\
f_{i,x}+ \frac{SL \times (t-T_{ss})}{T_{ds}} \qquad\qquad T_{ss} \leq t < T_{ds} 
\end{cases} ,
\label{eq:zmpEquation}
\end{equation}
\noindent
where $t$ denotes the time, which is reset at the end of each step ($t \geq T_{ss}+T_{ds}$), $T_{ss}$ , $T_{ds}$ are the duration of single and double support phases, respectively, $f_{i} = [f_{i,x} \quad f_{i,y}]$ is a set of planned foot positions on a 2D surface ($i \in \mathbb{N}$). After generating the ZMP trajectory, the trajectory of COM should be calculated which can be obtained by solving Equation~\ref{eq:lipm} as a boundary value problem. The boundary conditions that are considered to solve this differential equation are the positions of the COM at the beginning and at the end of a step. Accordingly, the trajectory of COM can be obtained using the following function:
\begin{equation}
\label{eq:com_traj_x0xf}
\resizebox{0.95\linewidth}{!} {$
	x(t) = p_x + \frac{ (p_x-x_f) \sinh\bigl(\omega\times(t - t_0)\bigl)+ (x_0 - p_x) \sinh\bigl(\omega\times(t - t_f)\bigl)}{\sinh(\omega\times(t_0 - t_f))}$ },
\end{equation}
\noindent
where $p_x$ represents the current ZMP position, $t_0$, $t_f$, $x_0$, $x_f$ are the times and corresponding positions of the COM at the beginning and at the end of a step, respectively. In this paper, we considered $T_{ds} = 0$ which means $p_x$ is constant during single support phase (center of current stance foot position) and it transits to the next step at the end of each step instantaneously~\cite{kasaei2018optimal}. In addition, $x_f$ is assumed to be between the current support foot and the next support foot ($\frac{f_{i} + f_{i+1}}{2}$). DCM reference trajectory can be generated by substituting the generated COM trajectory and its derivative into Equation~\ref{eq:dcm}. Although the mass of swing leg is not small, LIPM assumes a massless swing leg and does not consider the effect of swing leg motion. Thus, the swing leg trajectories should be smooth enough to reduce the effect of its motion. In our target, a cubic spline is used to generate the swing leg trajectories. 
\begin{figure}[!t]
	\label {exPlanner}
	\begin{centering}
		\begin{tabular}	{c c}			
			\includegraphics[width = 0.45\columnwidth, trim= 4cm 6cm 8cm 5cm,clip]{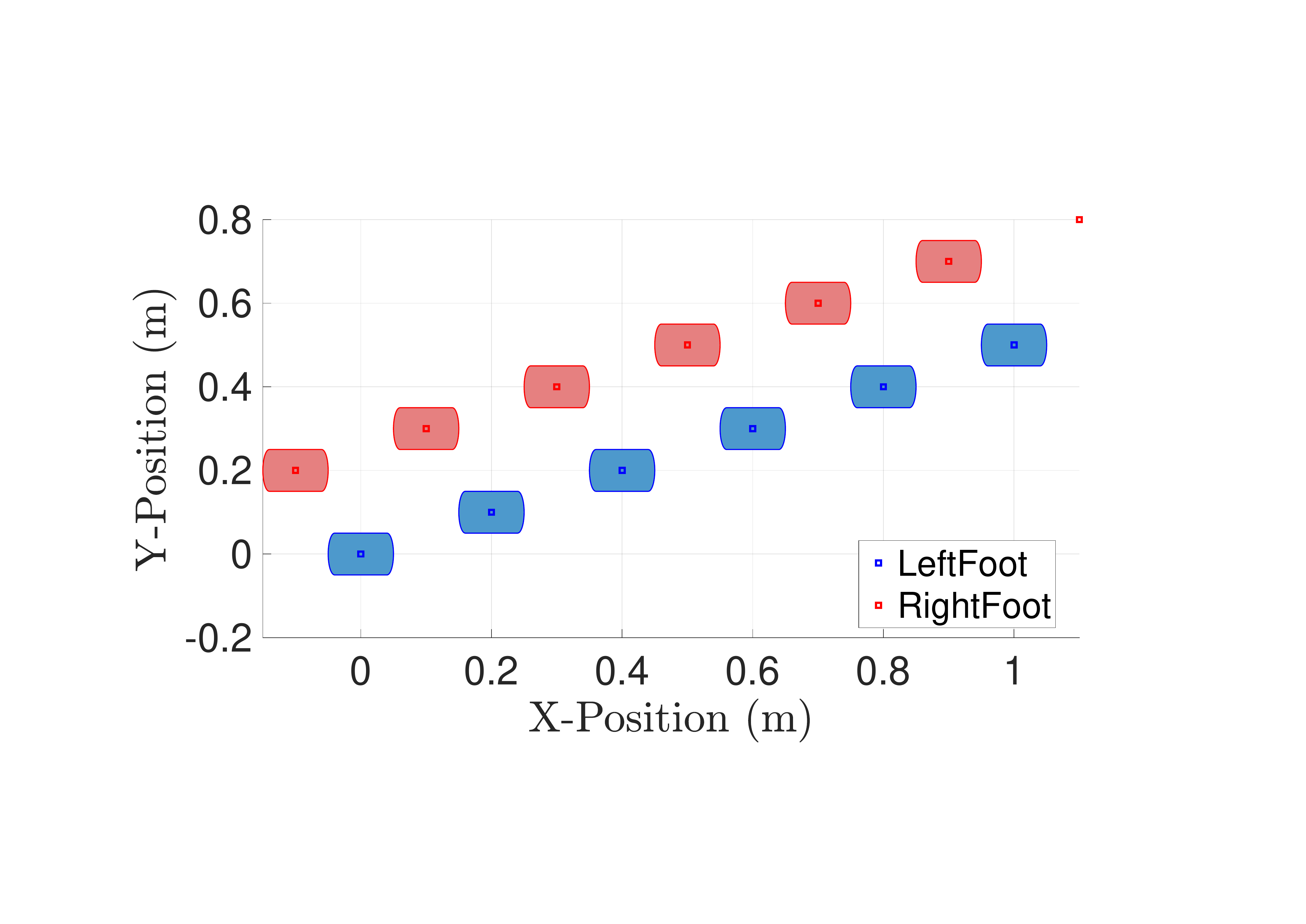}&
			\hspace{-2mm}\includegraphics[width = 0.5\columnwidth, trim= 4cm 6cm 6cm 6cm,clip]{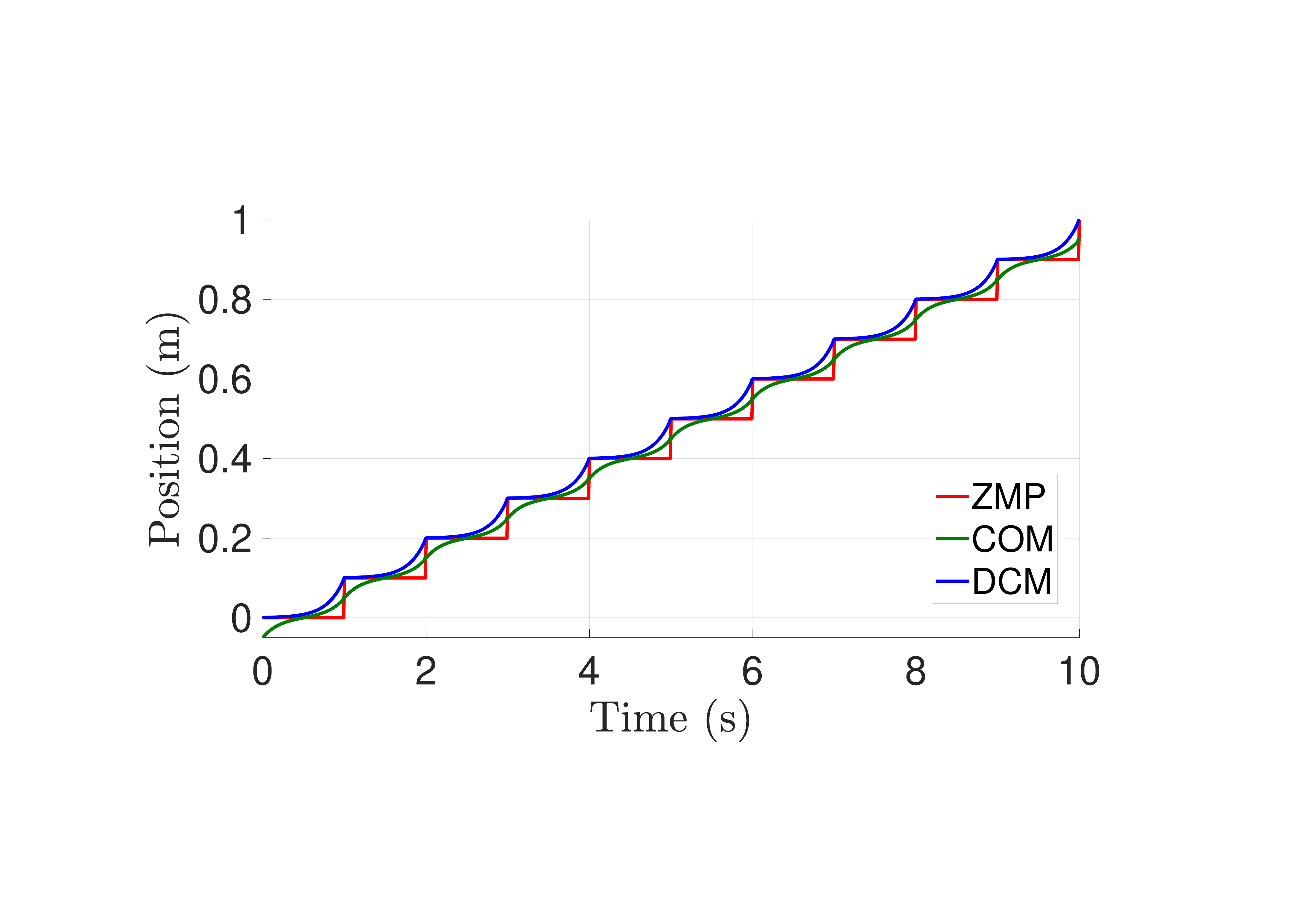}
		\end{tabular}
	\end{centering}
	\vspace{-0mm}
	\caption{ An exemplary trajectories for a ten-steps walking. Left: output of footstep planner , Right: corresponding ZMP, COM and DCM reference trajectories}
	\vspace{-5mm}
	\label{fig:exPlanner}
\end{figure}
To demonstrate the outputs of each planner, example trajectories have been planned for a ten-steps walking with a step duration~(SD) of 1$s$, step length~(SL) of 0.2$m$ and step width~(SW) of 0.1$m$. The generated plan is depicted in Fig.~\ref{fig:exPlanner}. Left plot of this figure represents the output of footstep planner and the right plot shows the corresponding ZMP, COM and DCM reference trajectories.

\section{Motion Tracking Control}
\label{sec:Control}
A feed-forward walking can be developed based on the generated reference trajectories but it is not stable in the presence of uncertainties. In this section, a Linear-Quadratic-Gaussian~(LQG) controller will be designed to track the desired reference trajectories robustly. Moreover, the robustness of the controller in the presence of process disturbances and also measurement noise will be shown.

\subsection{Overall architecture}
\begin{figure}[!t]
	\label {controller}
	\centering
	\includegraphics[width = 0.98\linewidth, trim= 0.1cm 8cm 1cm 3cm,clip]{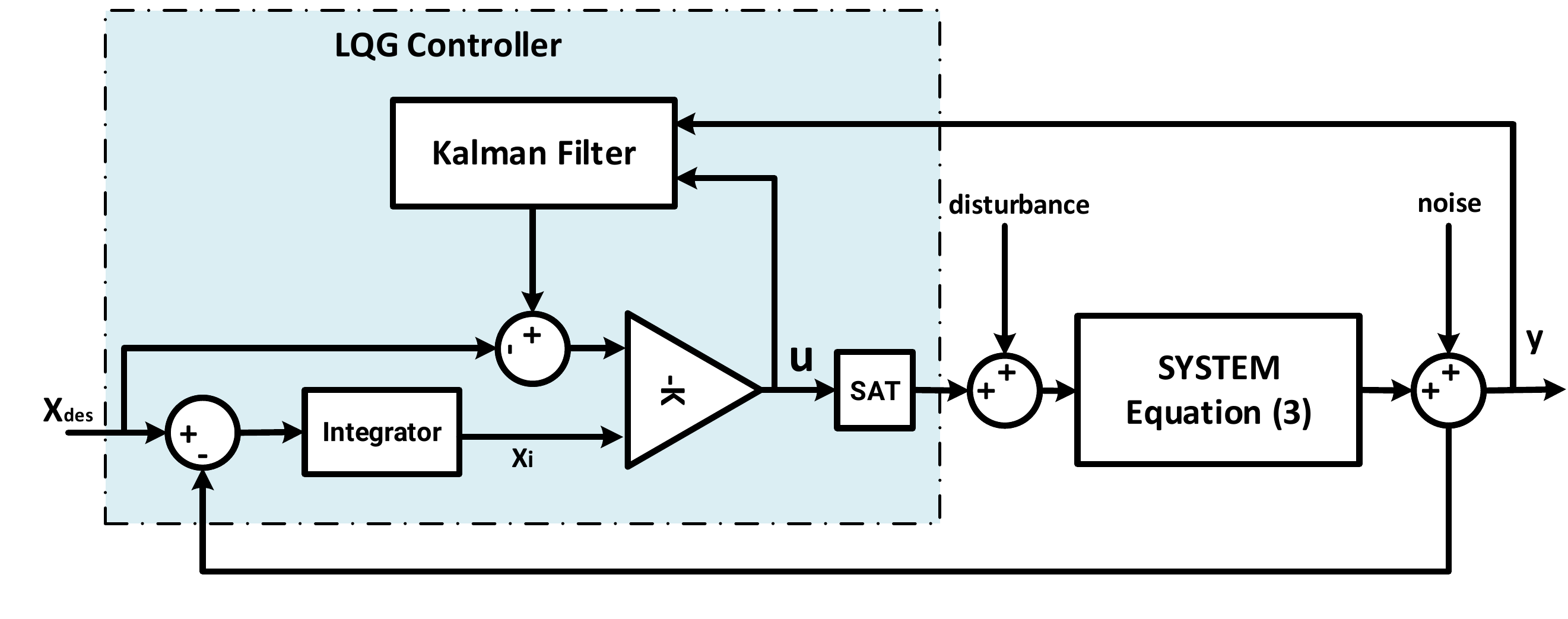}
	\vspace{-2mm}
	\caption{ The overall architecture of the controller.}
	\label{fig:controller}
	\vspace{-5mm}
\end{figure}

The overall architecture of the proposed controller is depicted in Fig.~\ref{fig:controller}. As shown in this figure, a Kalman filter is designed and used to estimate the states of the system in the presence of measurement as well as process noises. Moreover, according to the observability of the state's error in each control cycle, an integrator is used to eliminate steady-state error. Using the estimated state and output of the integrator, an optimal state-feedback control law for the tracking is designed as follows:
\begin{equation}
u = -K
\begin{bmatrix}
\tilde{x} - x_{des}\\
x_i
\end{bmatrix} \quad ,
\end{equation}
\noindent
where $\tilde{x}$, $x_{des}$ represent estimated states and desired states, respectively. $x_i$ is the integrator output, $K$ represents the optimal gain matrix which is designed to minimize the following cost function:
\begin{equation}
J(u) = \int_{0}^{\infty} \{ z^\intercal Q z + u^\intercal R u \} dt \quad ,
\end{equation}
\noindent
where $z = [\tilde{x} \quad x_i]^\intercal $, $Q$ and $R$ are a tradeoff between
tracking performance and cost of control effort. There is a straightforward solution to find $K$ by solving a differential equation which is called the Riccati Differential Equation (RDE). The quality of this controller depends on the choice of Q as well as R, and they generally are selected using some trial and error. 
\subsection{Robustness analysis}
The controller is the most important part of a real humanoid robot due to the unstable nature of this type of robot. In a real robot, measurements are never perfect and always are affected by unknown noises. A robust controller should guarantee the tracking performance in presence of measurement and process noises. 
The remainder of this section is dedicated to analyzing the robustness of the proposed controller in the presence of uncertainties using a simulated robot. The mass of the simulated robot is 30Kg, the height of COM is considered to be 1m and the foot length is 0.2m.
\begin{figure}[!t]
	\label {controller_sim_res}
	\begin{centering}
		\begin{tabular}	{c c}			
			\includegraphics[width = 0.48\columnwidth, trim= 4.5cm 6cm 6.5cm 7cm,clip] {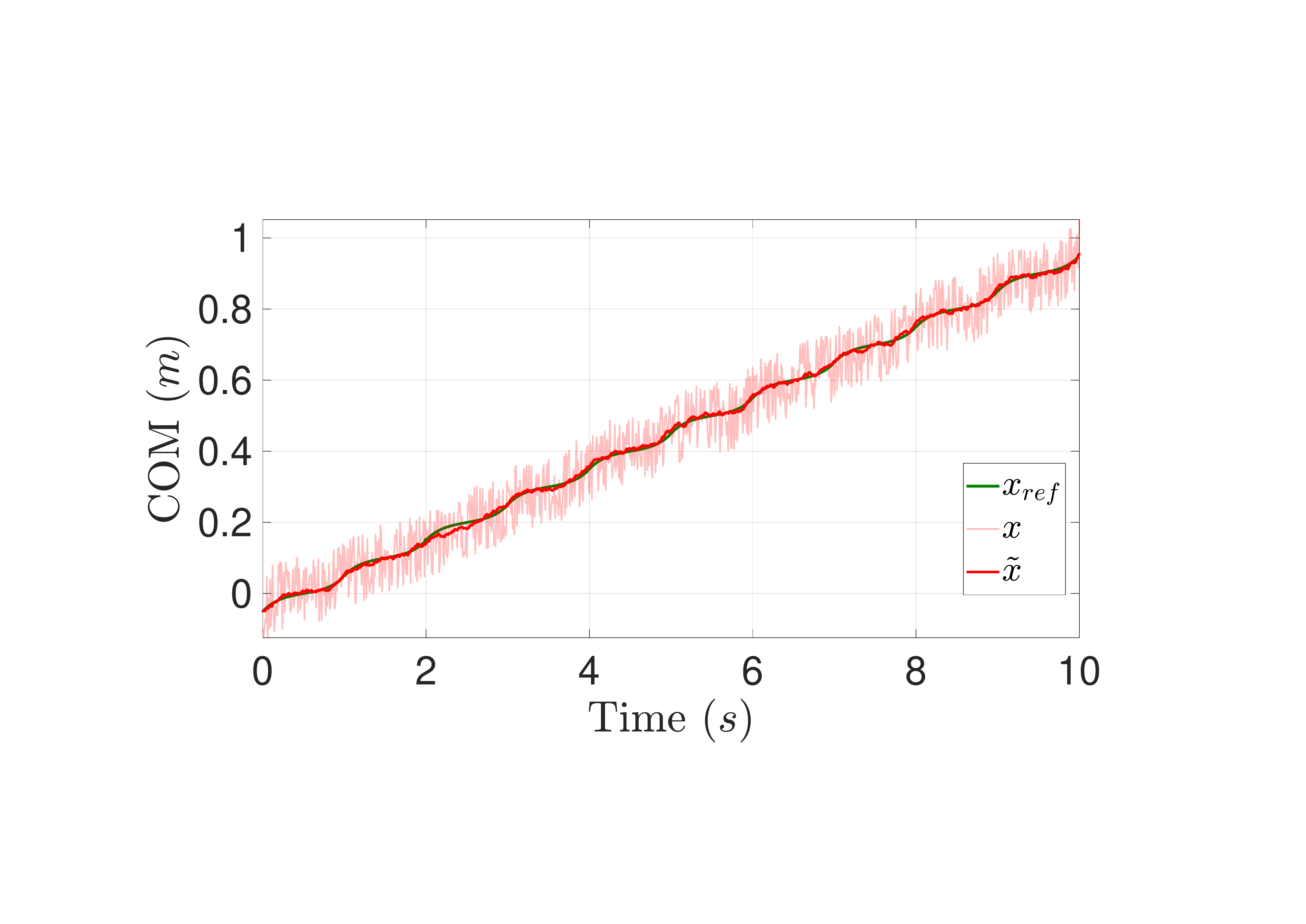}&  
			\includegraphics[width = 0.48\columnwidth, trim= 4.5cm 6cm 6.5cm 7cm,clip] {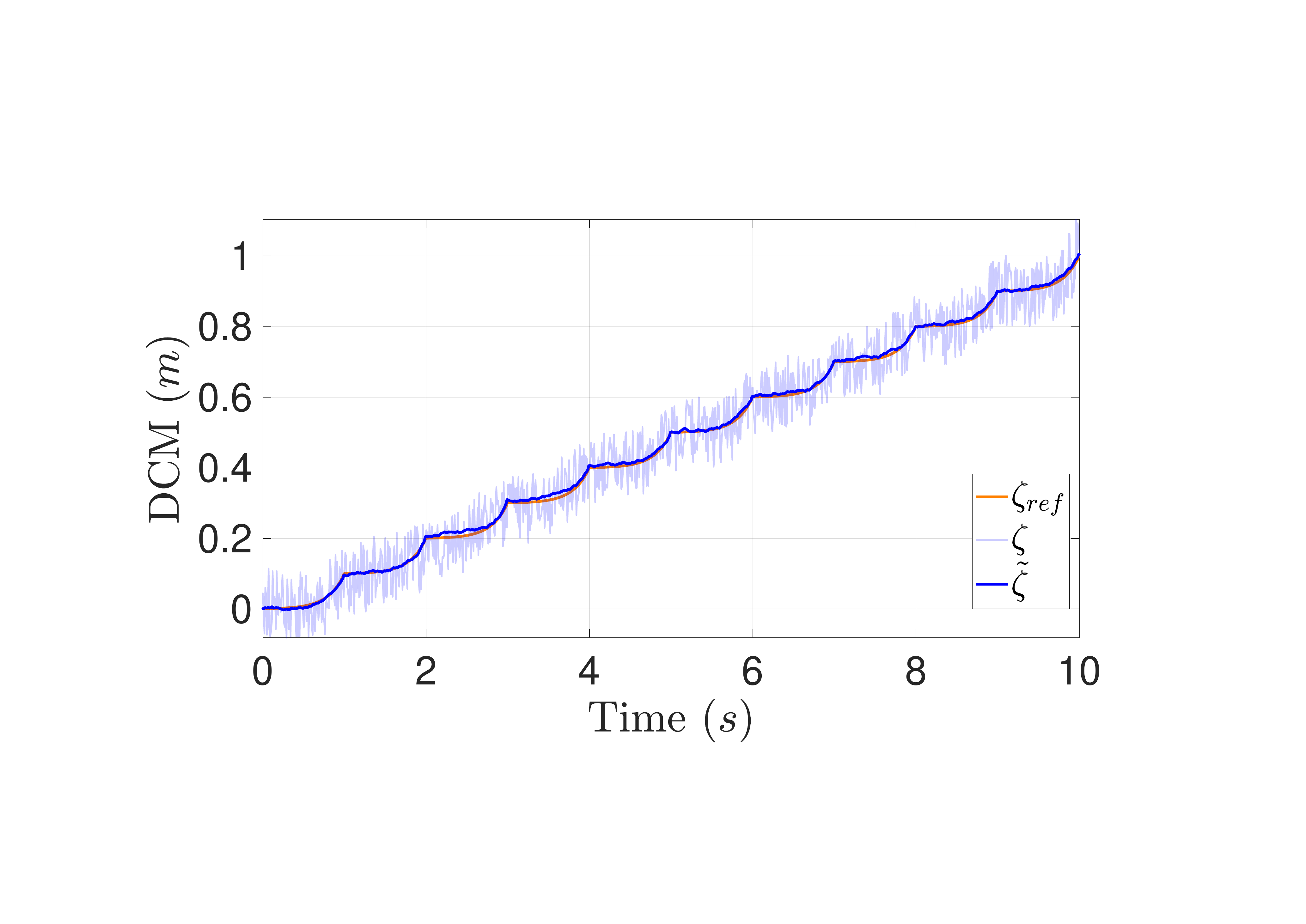}
		\end{tabular}
	\end{centering}
	\vspace{-2mm}
	\caption{ Simulation results of examining the robustness w.r.t. measurement errors. The measurements are modeled as a stochastic process by adding two independent Gaussian noises \mbox{($-0.05\le v \le 0.05 $)} to the system output.}
	\vspace{-1mm}
	\label{fig:controller_sim_res}
\end{figure}
\begin{figure}[!t]
	\label {controller_sim_ext_push_res}
	\begin{centering}
		\hspace{-3mm}
		\begin{tabular}	{c c c}			
			F = 45 N & F = 70 N & F = 90 N \\
			\includegraphics[width = 0.32\linewidth, trim= 4.8cm 8.8cm 7.1cm 6.7cm,clip] {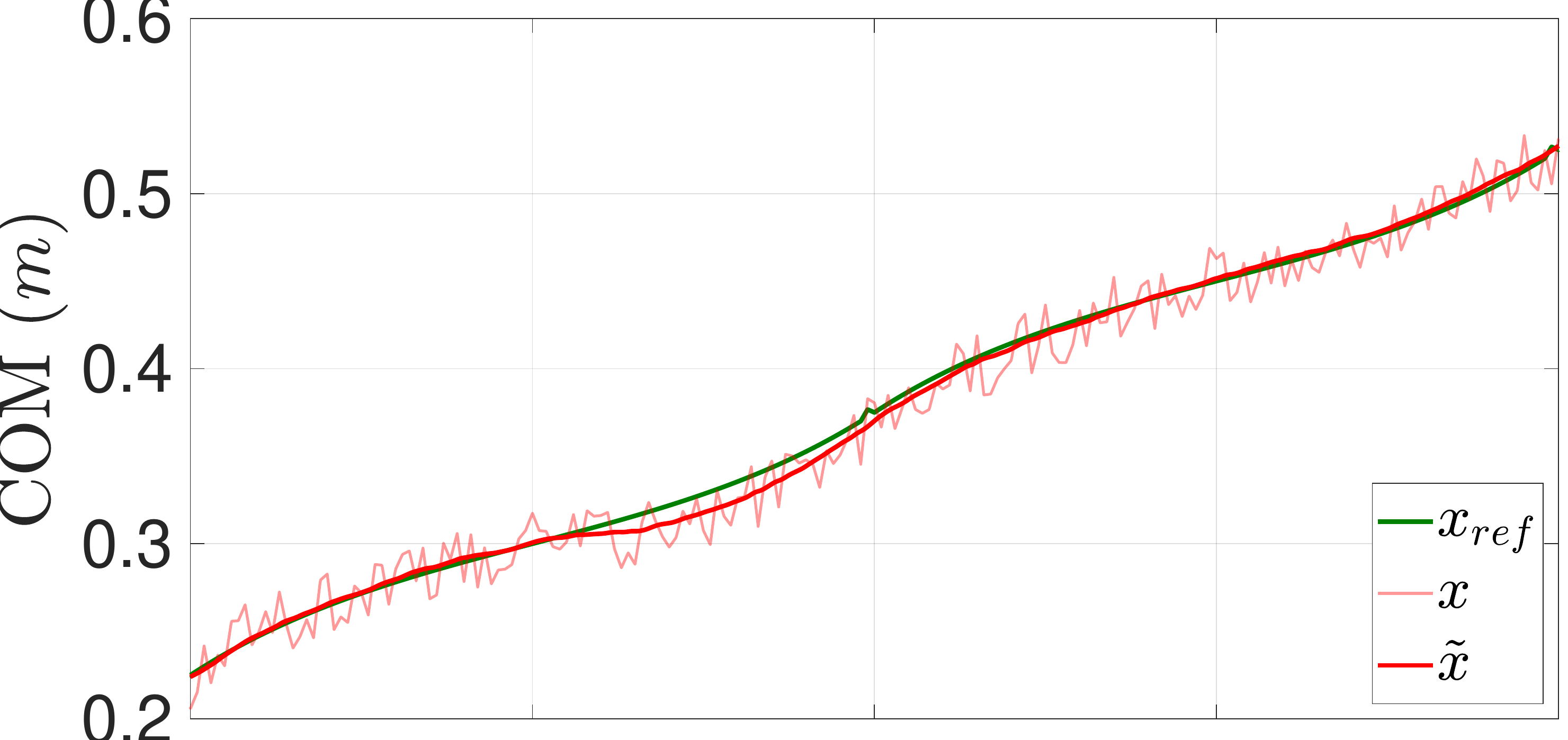} &
			\includegraphics[width = 0.29\linewidth, trim= 8.1cm 8.8cm 7.1cm 7cm,clip] {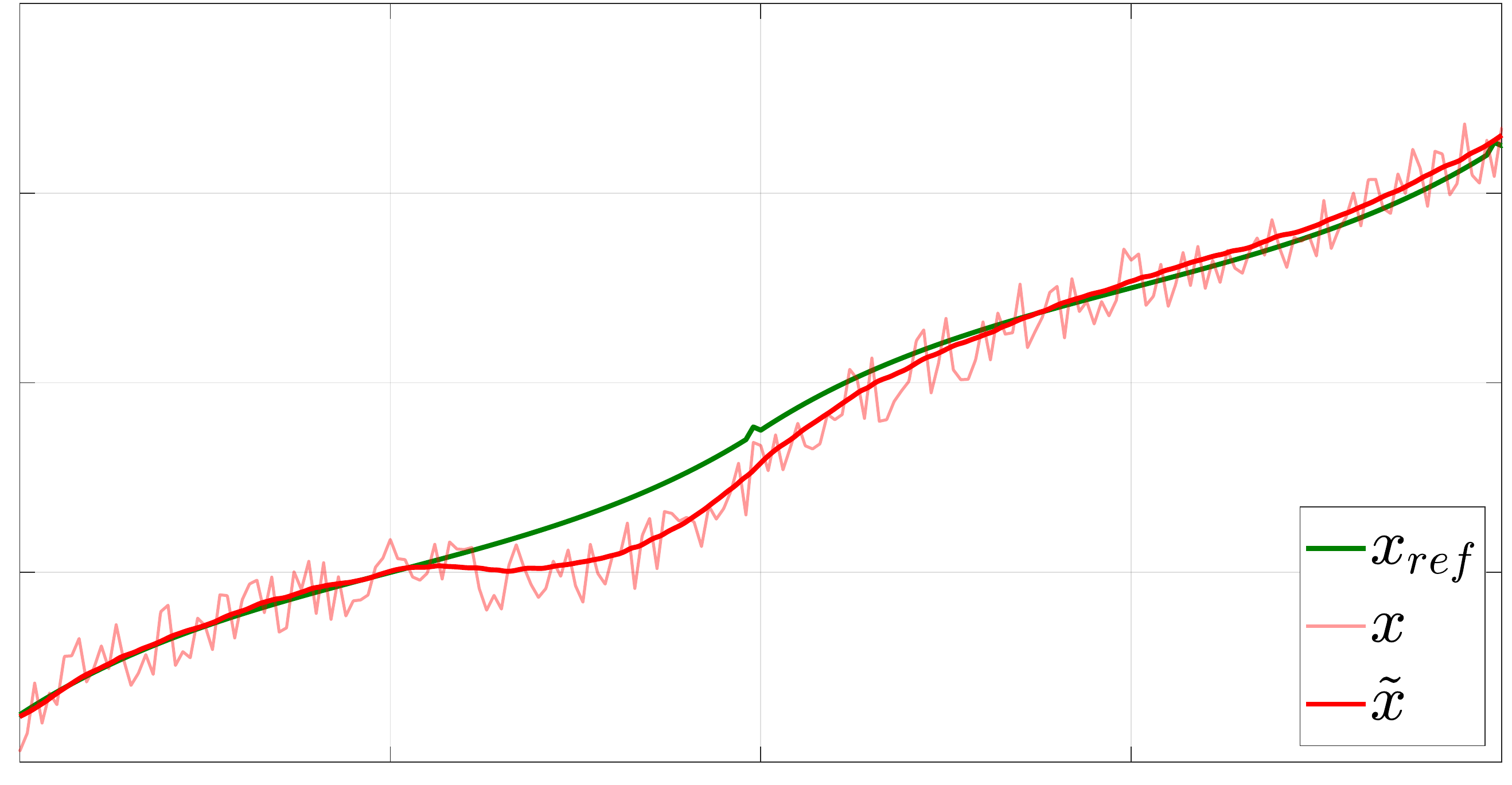} &
			\includegraphics[width = 0.29\linewidth, trim= 8.1cm 8.8cm 7.1cm 7cm,clip] {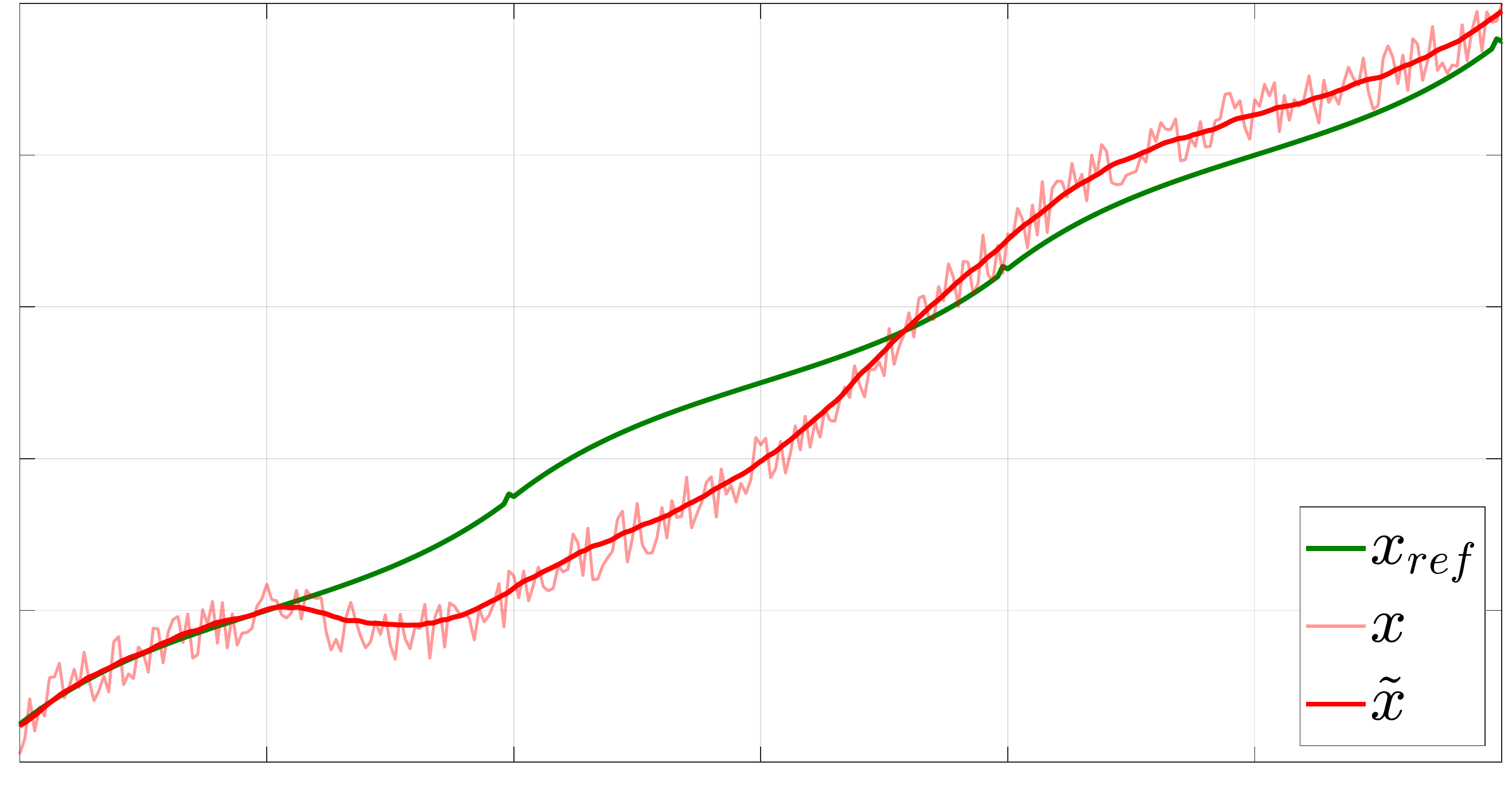} \\ 
			
			\includegraphics[width = 0.32\linewidth, trim= 4.8cm 8.8cm 7.1cm 6.7cm,clip] {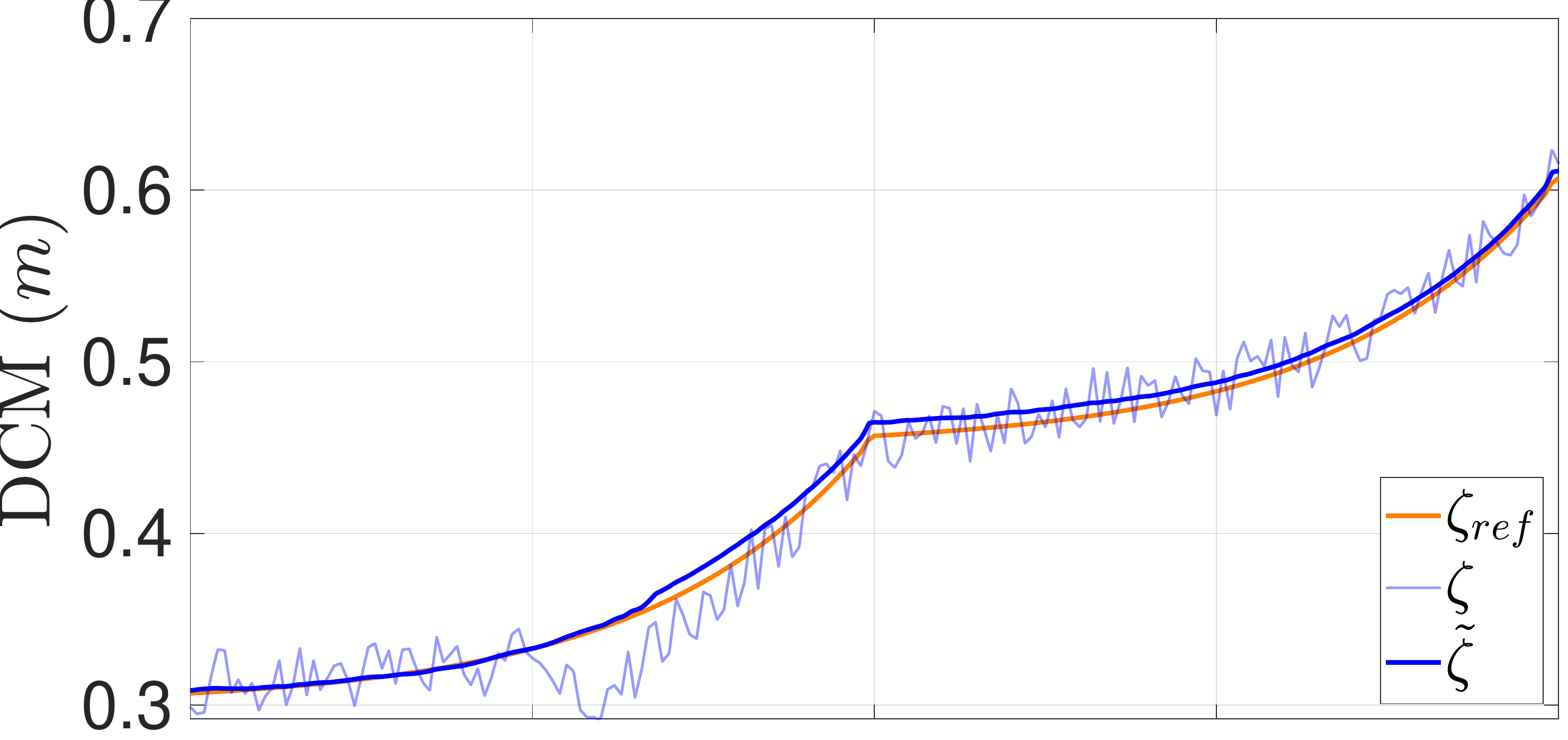} &
			\includegraphics[width = 0.29\linewidth, trim= 8.1cm 8.8cm 7.1cm 7cm,clip] {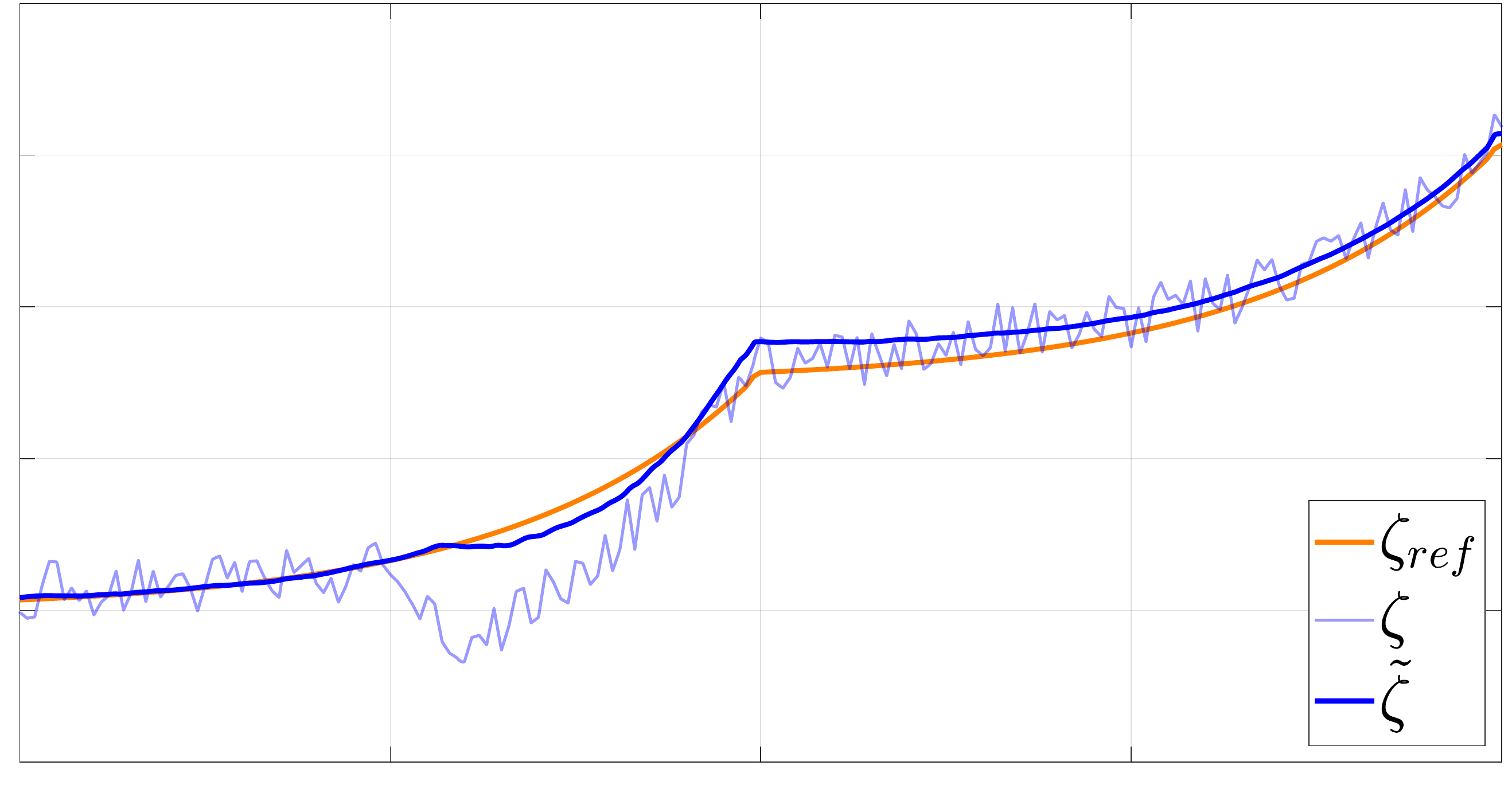} &
			\includegraphics[width = 0.29\linewidth, trim= 8.1cm 8.8cm 7.1cm 7cm,clip] {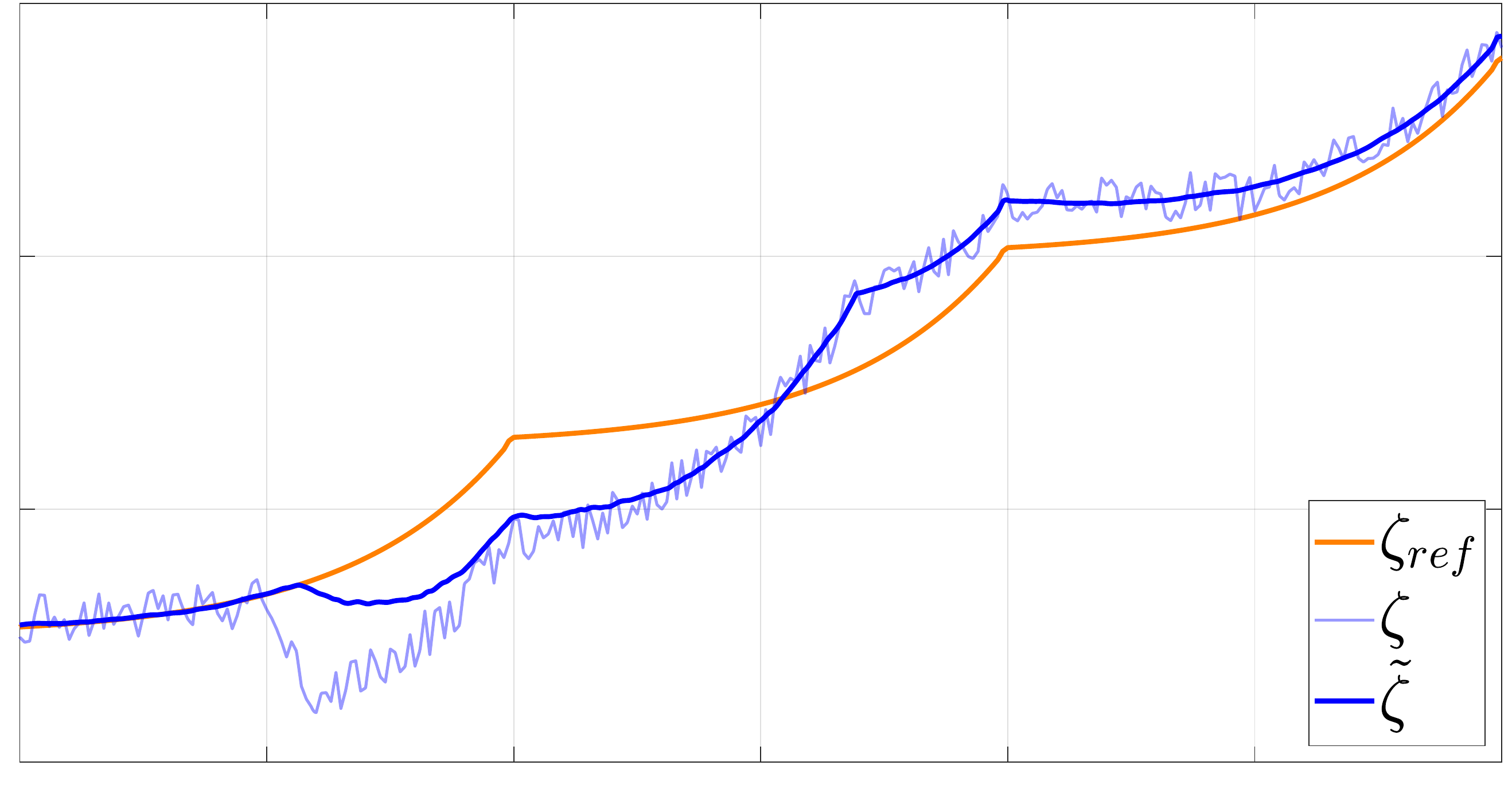} \\ 
			
			\includegraphics[width = 0.32\linewidth, trim= 4.8cm 6cm 7.1cm 6.7cm,clip] {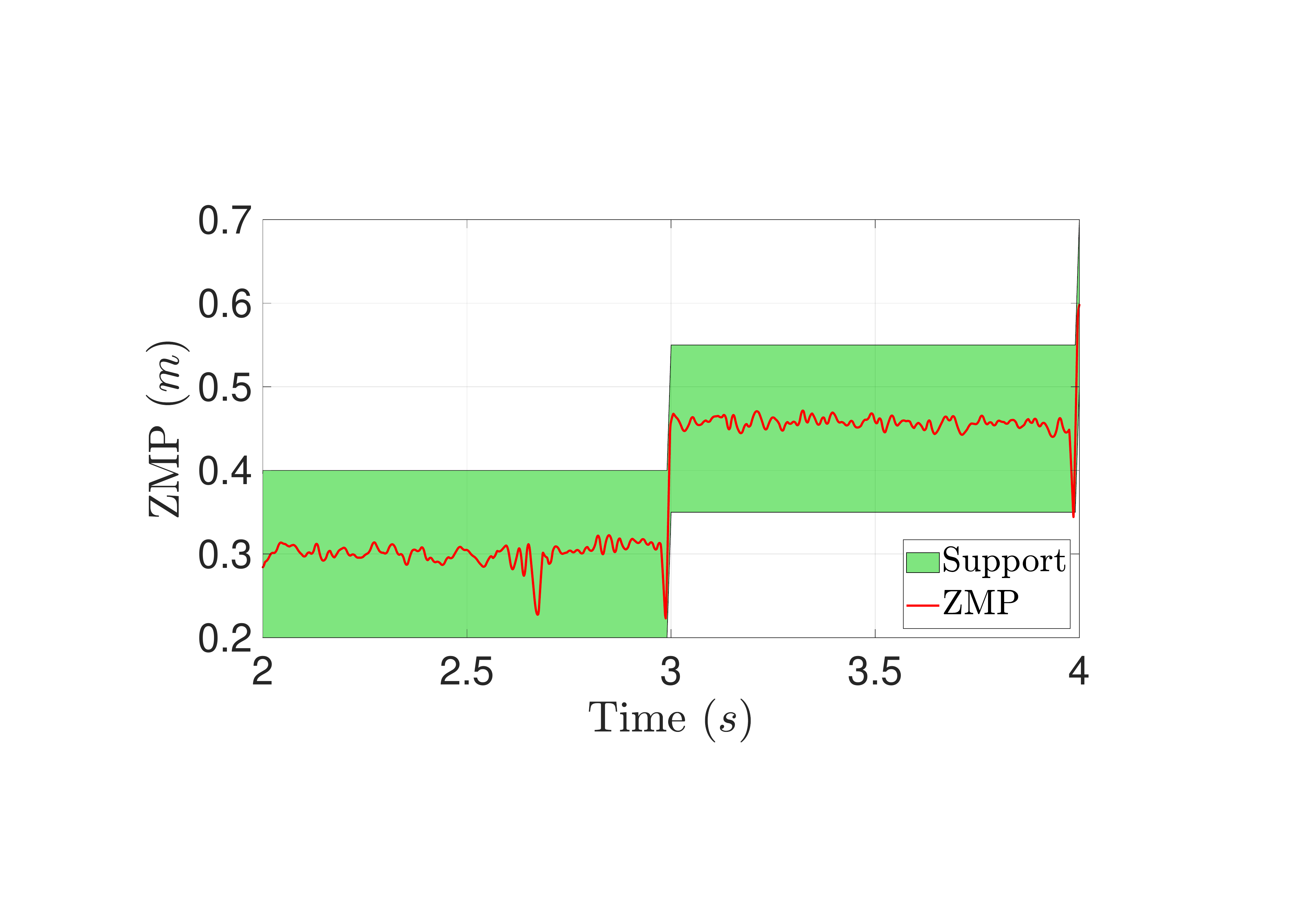} &
			\includegraphics[width = 0.29\linewidth, trim= 8.1cm 6cm 7.1cm 7cm,clip] {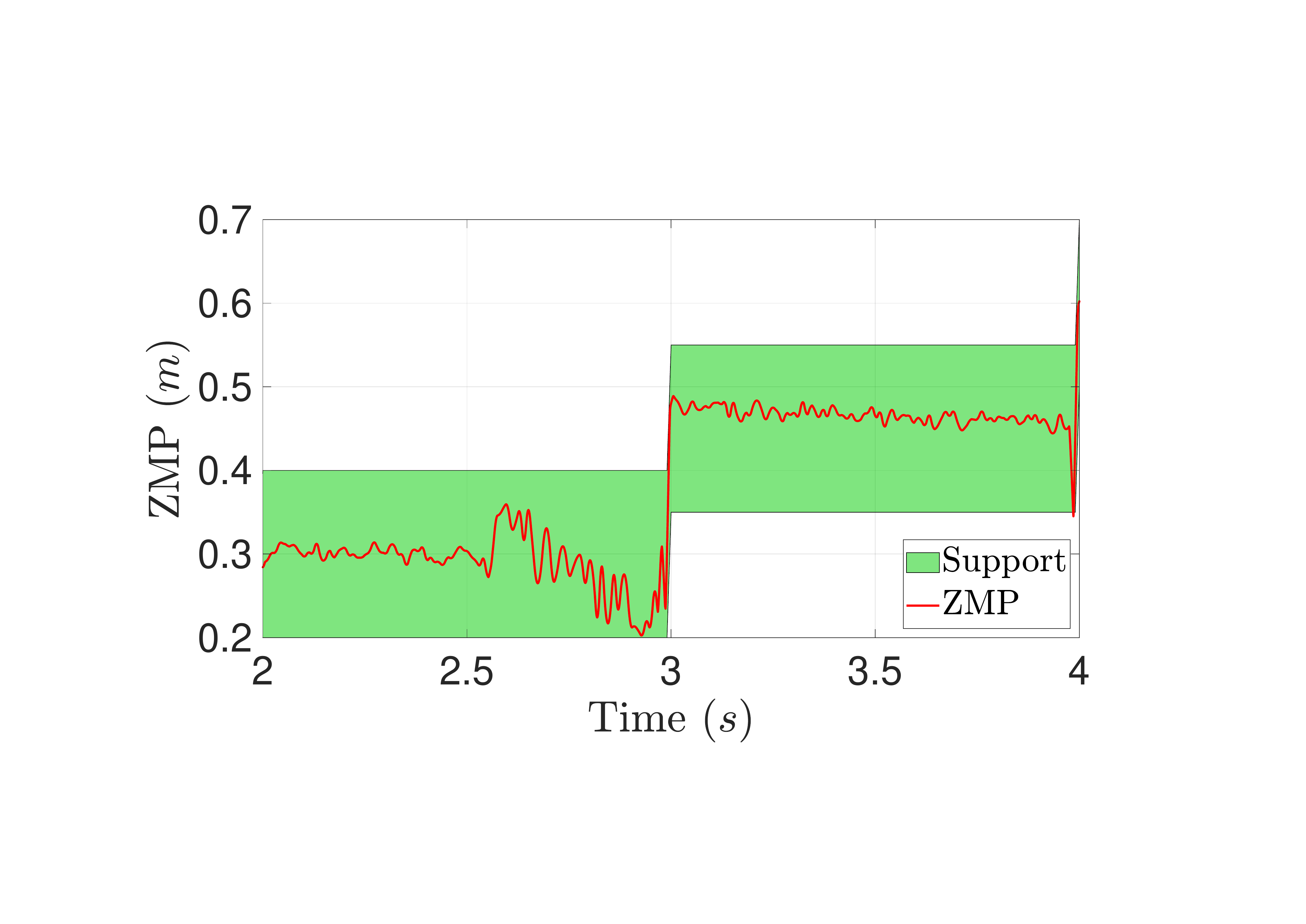} &
			\includegraphics[width = 0.29\linewidth, trim= 8.1cm 6cm 7.1cm 7cm,clip] {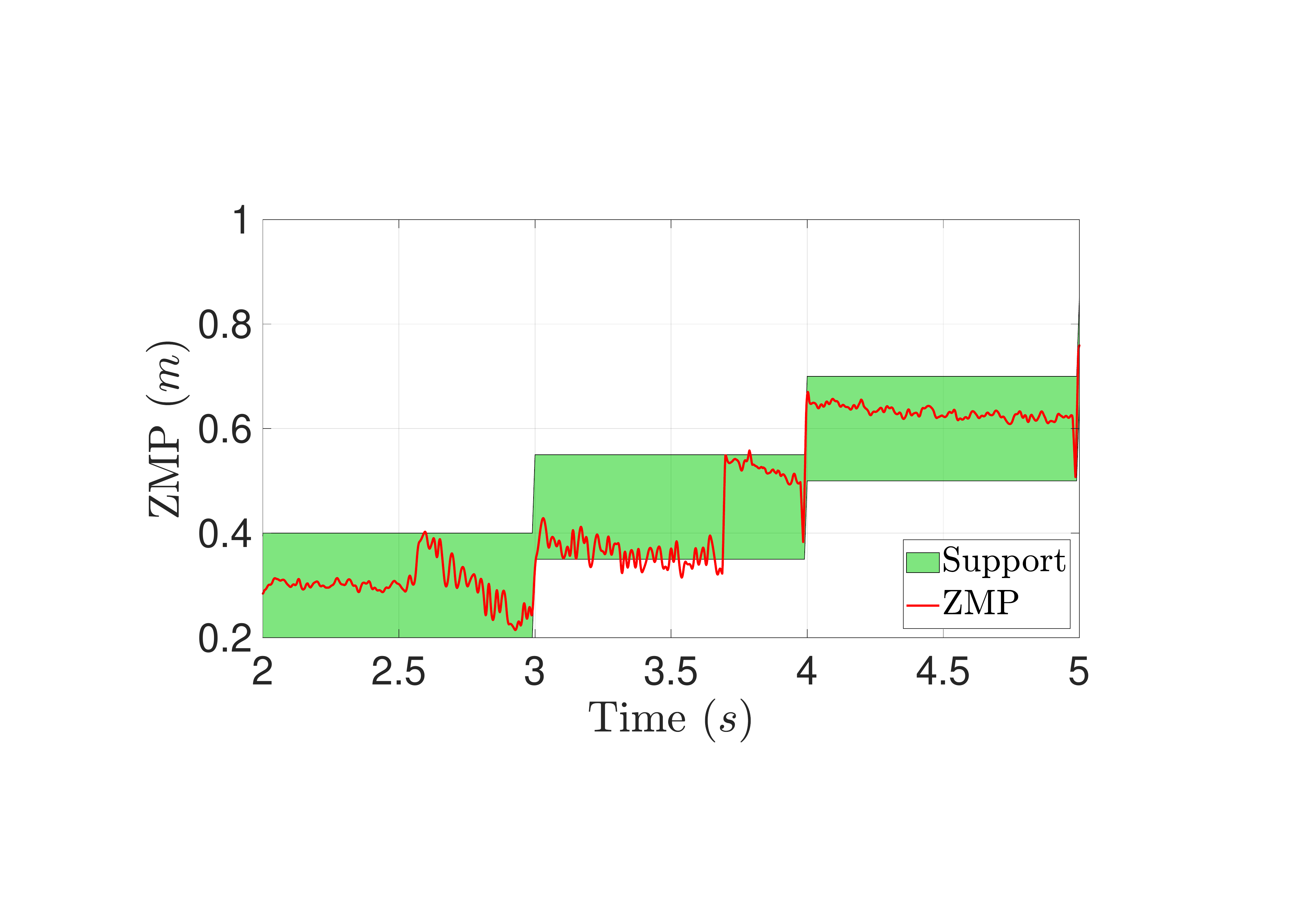} \\ 
			\vspace{-4mm}\\ F = -45 N & F = -70 N & F = -90 N \\									
			\includegraphics[width = 0.32\linewidth, trim= 4.8cm 8.8cm 7.25cm 6.7cm,clip] {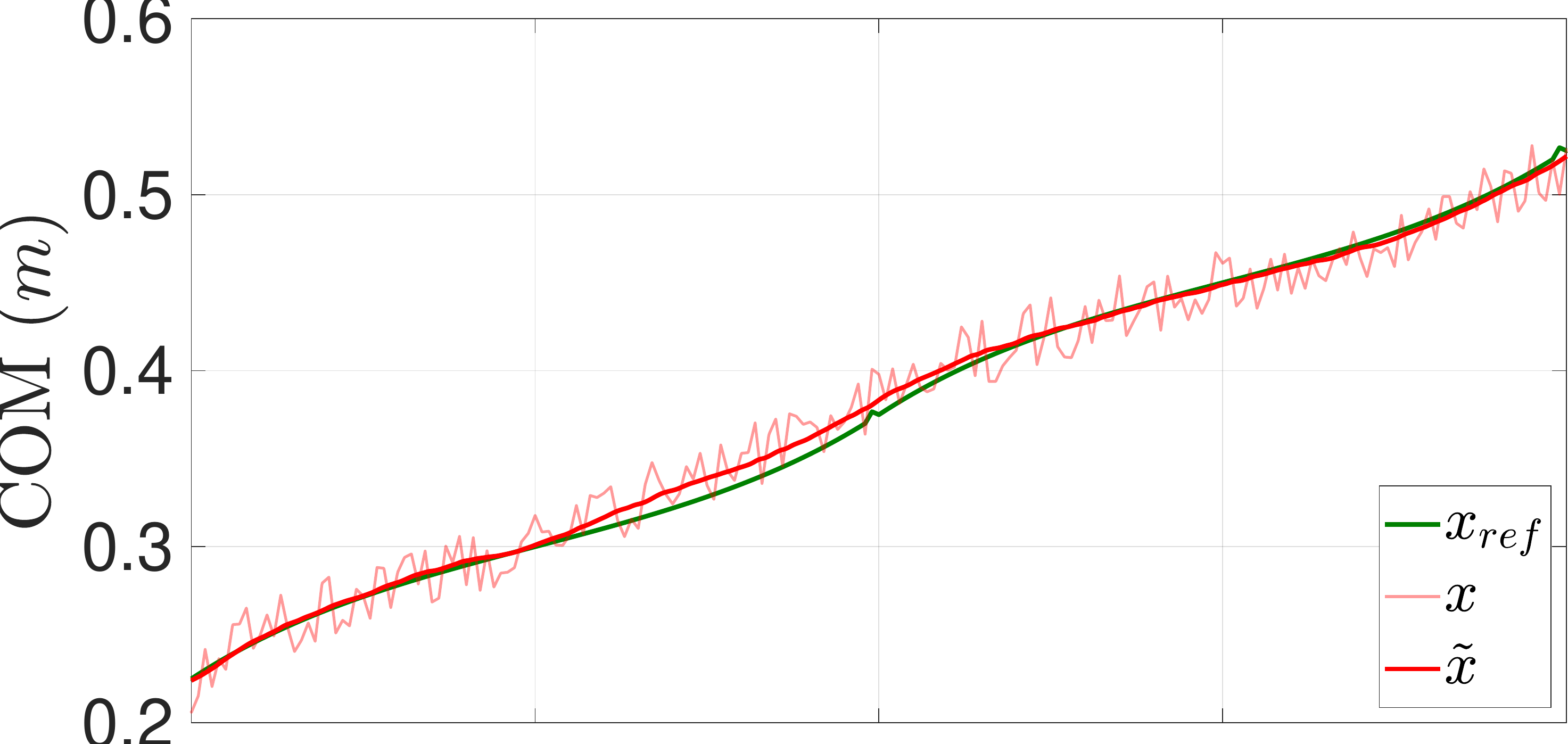} &
			\includegraphics[width = 0.29\linewidth, trim= 8.1cm 8.8cm 7.25cm 7cm,clip] {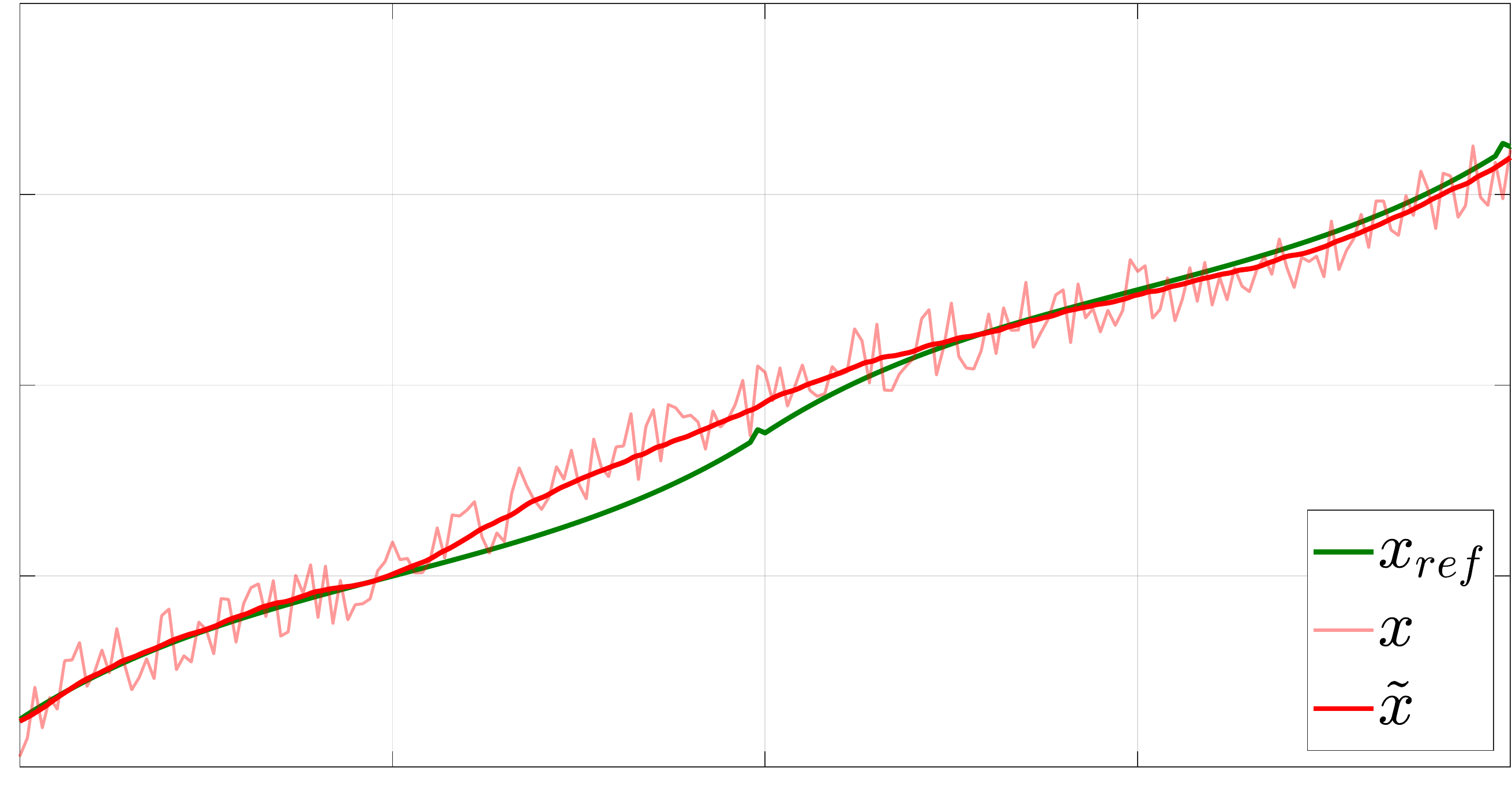} &
			\includegraphics[width = 0.29\linewidth, trim= 8.1cm 8.8cm 7.1cm 7cm,clip] {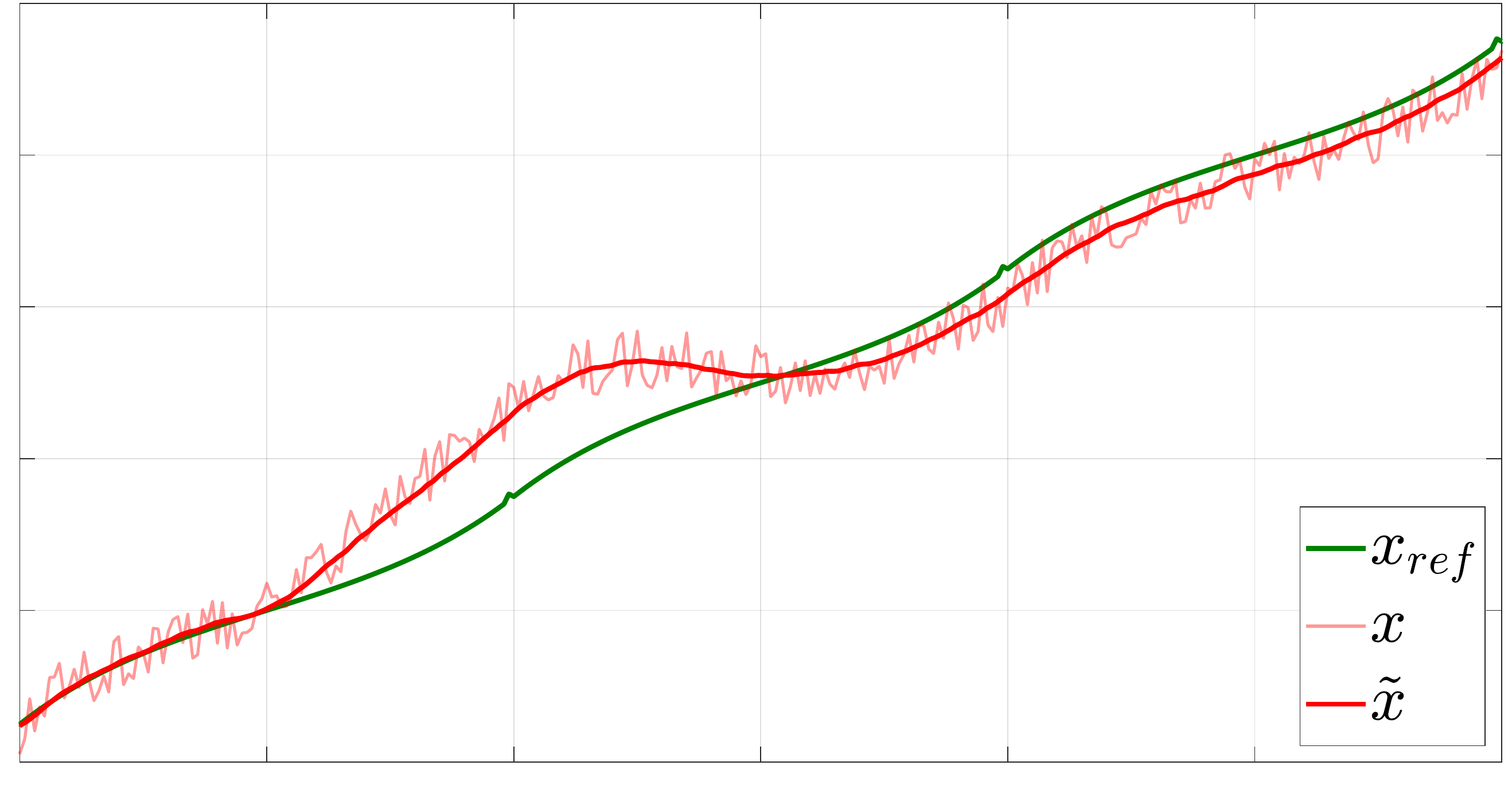} \\ 
			
			\includegraphics[width = 0.32\linewidth, trim= 4.8cm 8.8cm 7.25cm 6.7cm,clip] {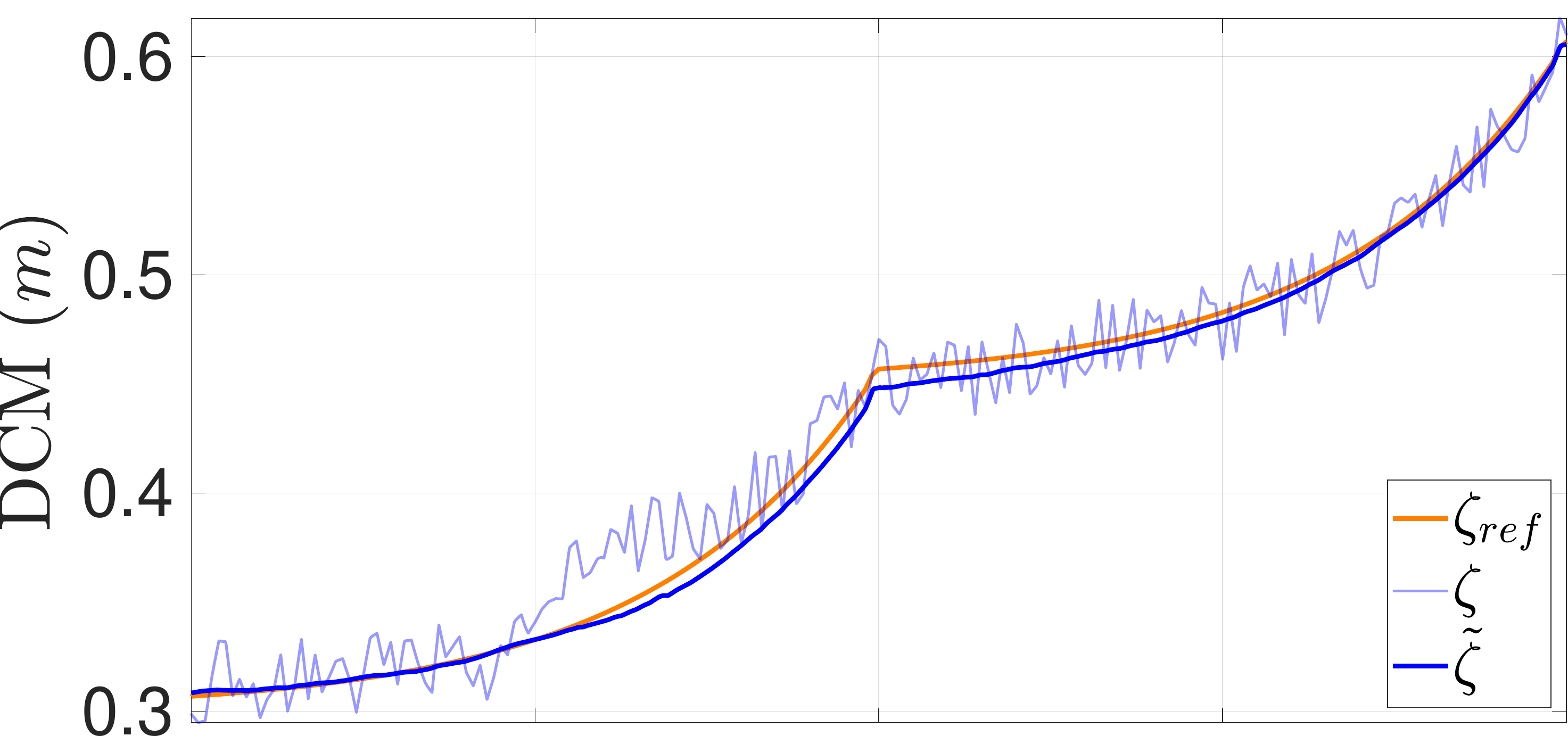} &
			\includegraphics[width = 0.29\linewidth, trim= 8.1cm 8.8cm 7.25cm 7cm,clip] {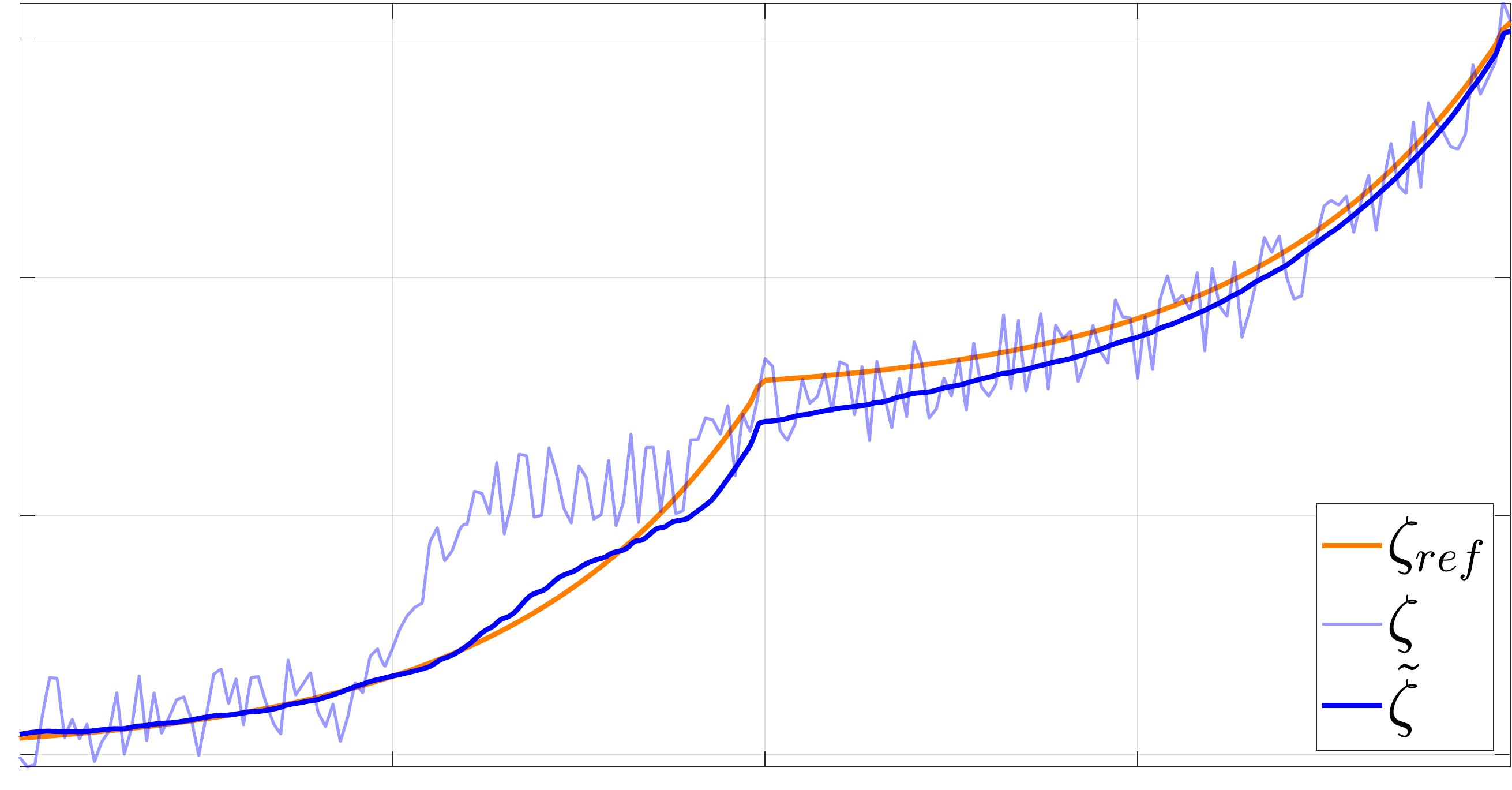} &
			\includegraphics[width = 0.29\linewidth, trim= 8.1cm 8.8cm 7.1cm 7cm,clip] {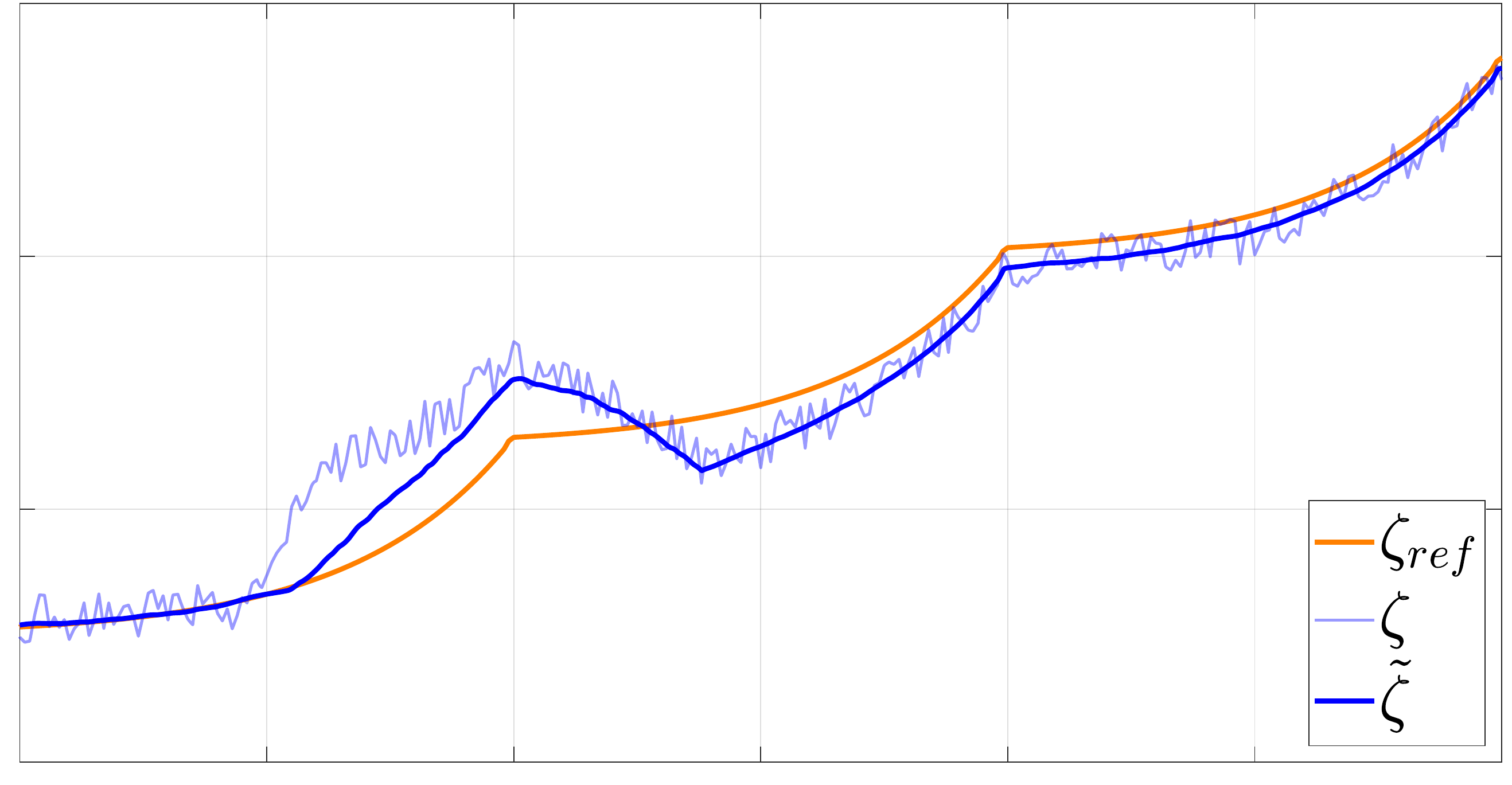} \\ 
			
			\includegraphics[width = 0.32\linewidth, trim= 4.8cm 6cm 7.25cm 6.7cm,clip] 	{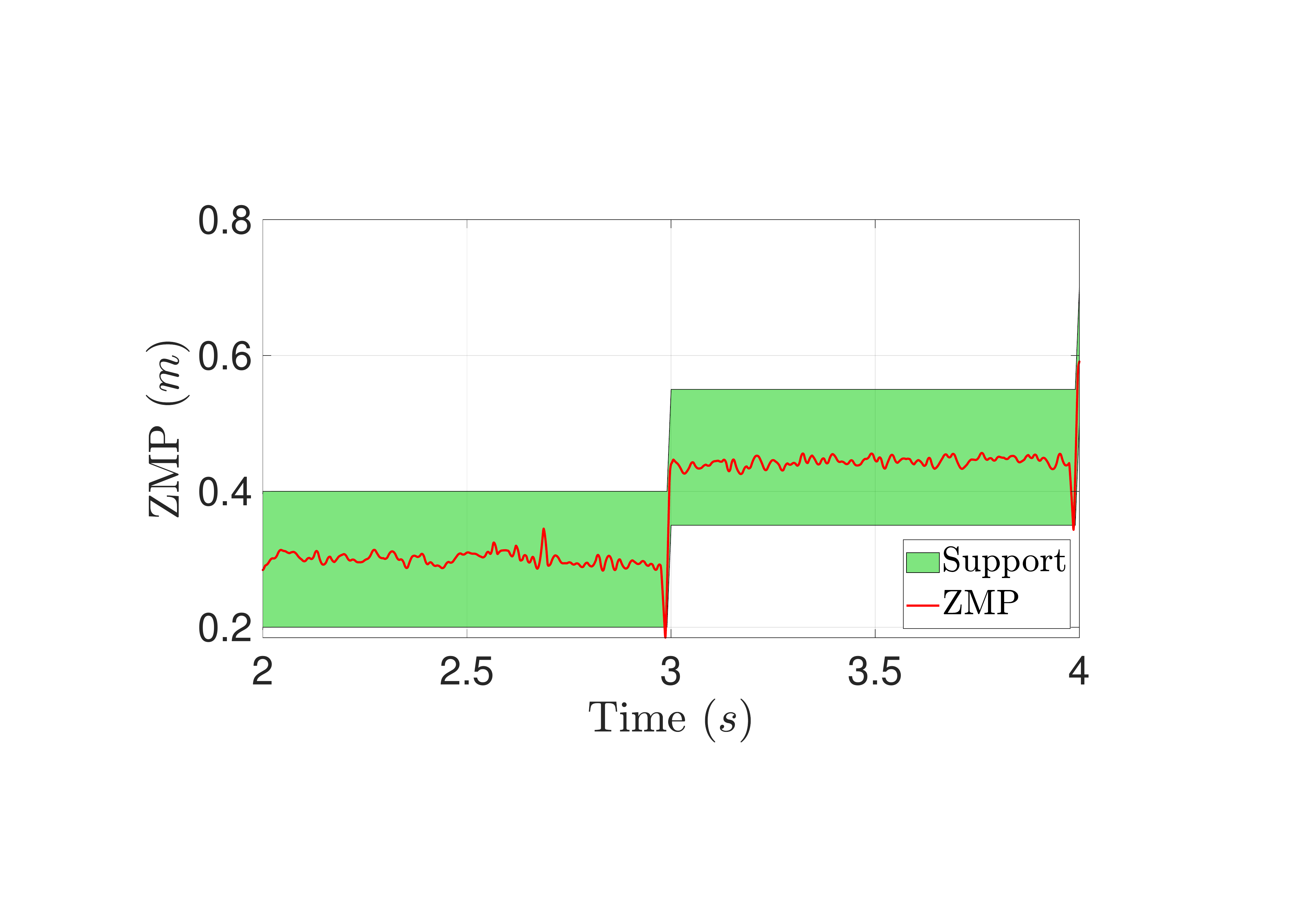} &
			\includegraphics[width = 0.29\linewidth, trim= 8.1cm 6cm 7.25cm 7cm,clip] {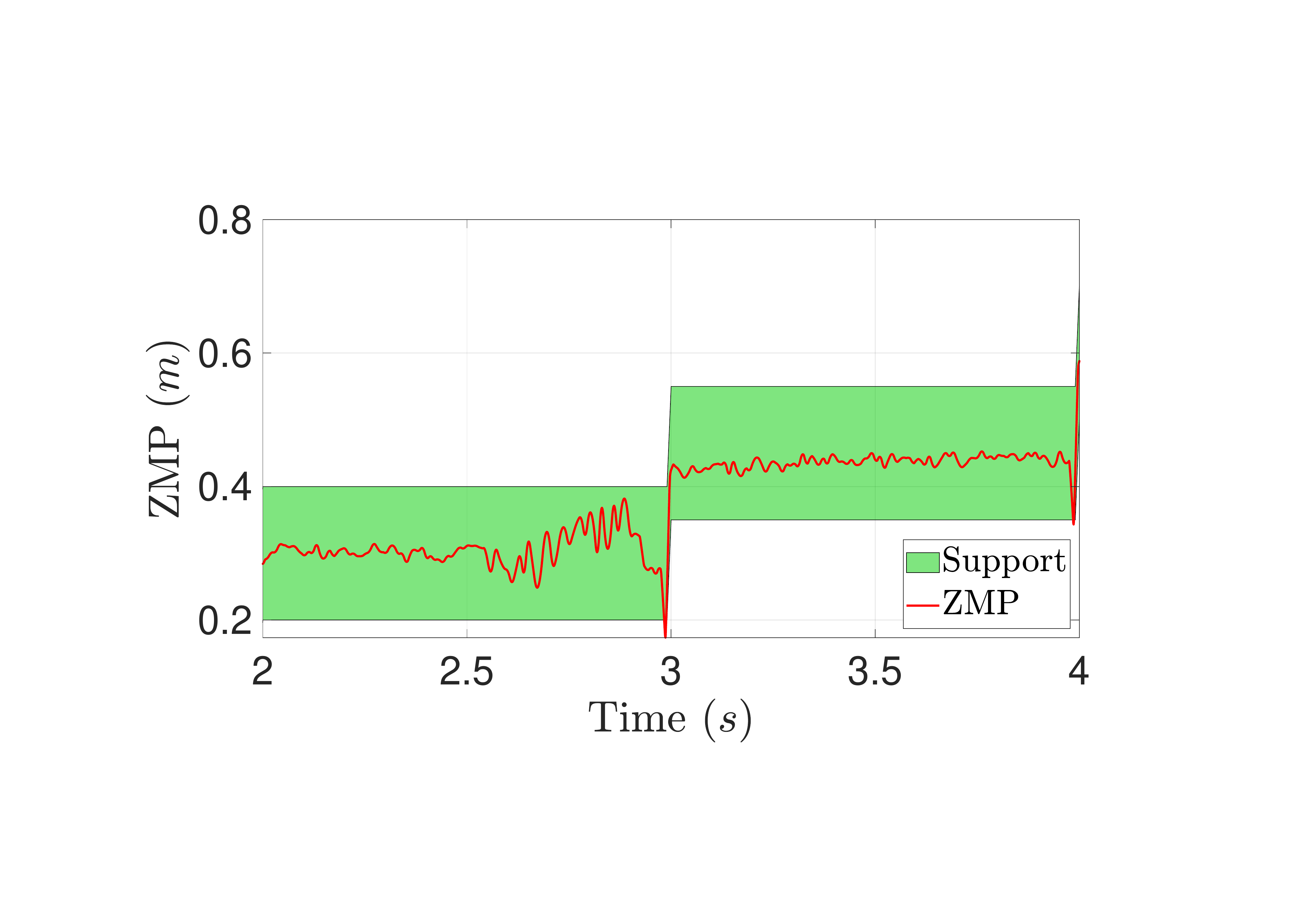} &
			\includegraphics[width = 0.29\linewidth, trim= 8.1cm 6cm 7.1cm 7cm,clip] {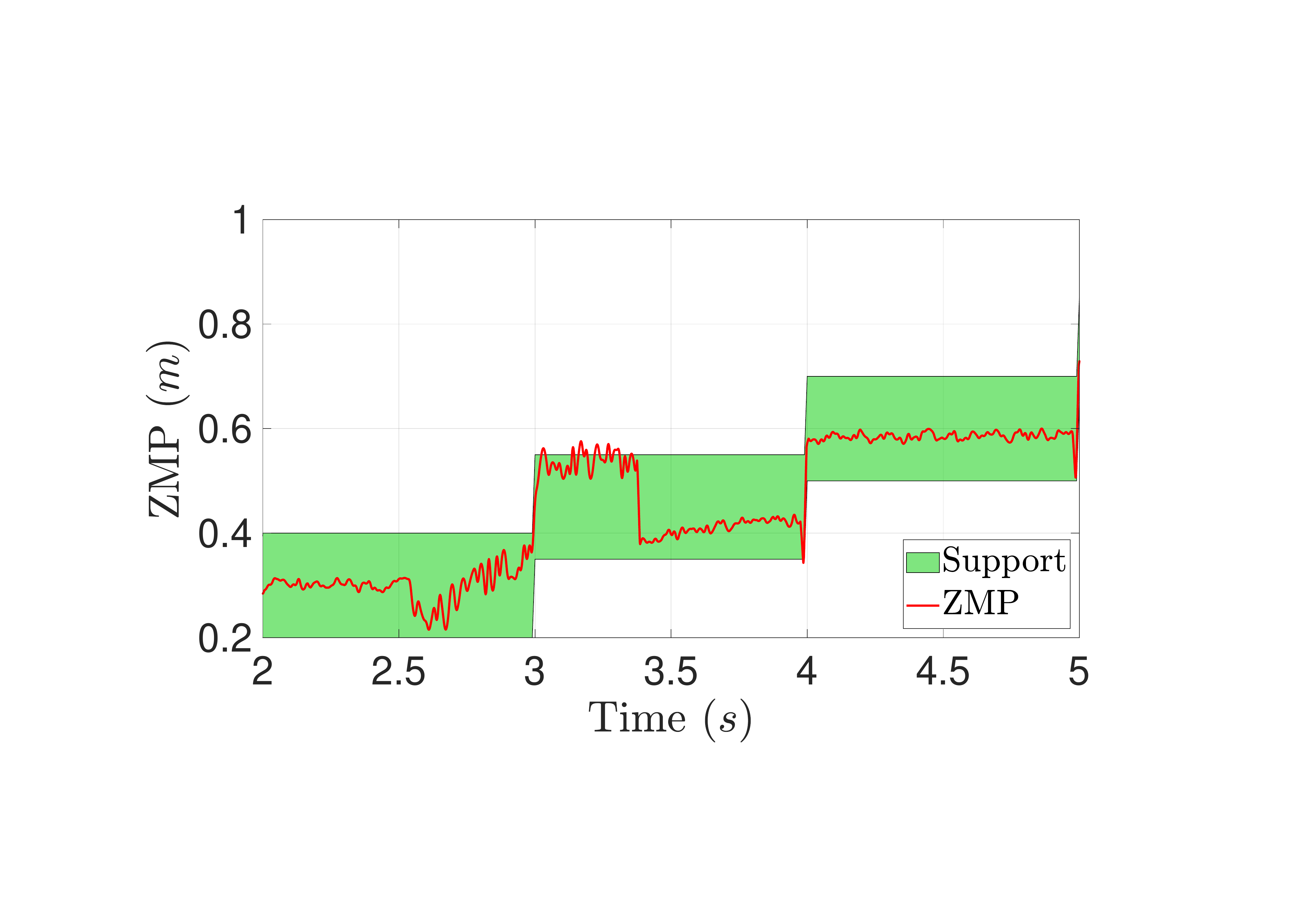} \\ 
		\end{tabular}
	\end{centering}
	\vspace{-2mm}
	\caption{ Simulation results of examining the robustness w.r.t. external disturbance. Each column represents the simulation results of forward and backward push with specified force.}
	\vspace{-5mm}
	\label{fig:controller_sim_ext_push_res}
\end{figure}

\textbf{Robustness w.r.t. measurement errors:} to examine this robustness, a simulation has been carried out. In this simulation, two independent zero-mean Gaussian noises are added to our measurements in order to simulate a noisy situation which occurs because of some reasons like noisy outputs of the sensors and also using inaccurate dynamics model. In this simulation, the generated trajectories of ten-steps walking (see Fig.~\ref{fig:exPlanner}) are used as desired trajectories and the controller should track them in presence of unknown but bounded noises. The results of this simulation are shown in Fig.~\ref{fig:controller_sim_res}. As shown in this figure, the controller is robust regarding measurements noise and could track the references well.

\textbf{Robustness w.r.t. unknown external disturbance:} In some situations like conflicting with an obstacle or being pushed by someone or other situations like these, an unknown external force will be applied to the robot. A robust controller should be able to react against these types of disturbances and keep the stability of the robot. To analysis this robustness, a set of simulations has been carried out to check the withstanding level of the controller against these disturbances. In these simulations, while the robot is walking (ten-steps walking scenario), a disturbance will be applied to its COM at $t= 2.5s$ and the impact duration is $ \Delta t = 10 ms$. This simulation has been repeated six times with different directions and amplitude of disturbances. The simulation results are depicted in Fig.~\ref{fig:controller_sim_ext_push_res}. To determine the maximum level of withstanding of the controller, the amplitude of the impact has been increased until the robot could not track the references and fell. Based on these simulation results, $F=91.4N$, $F=-97.2N$ were the maximum level of withstanding of the controller.
\begin{figure}[!t]
	\label {controller_diff_h}
	\begin{centering}
		\begin{tabular}	{c c}			
			\includegraphics[width = 0.45\columnwidth, trim= 4.5cm 8.8cm 6.5cm 6.5cm,clip] {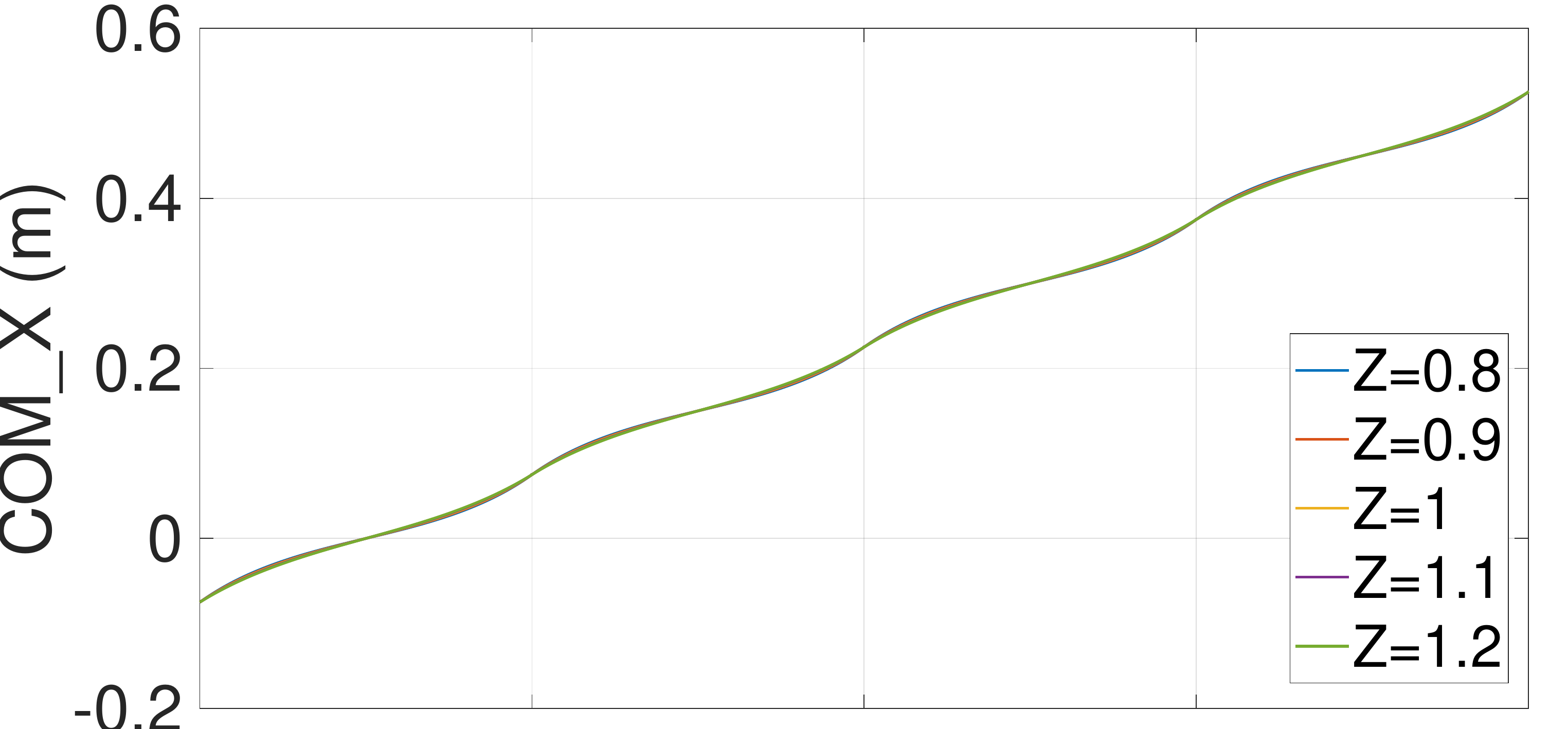}& 
			\includegraphics[width = 0.45\columnwidth, trim= 4.5cm 8.8cm 6.5cm 6.5cm,clip] {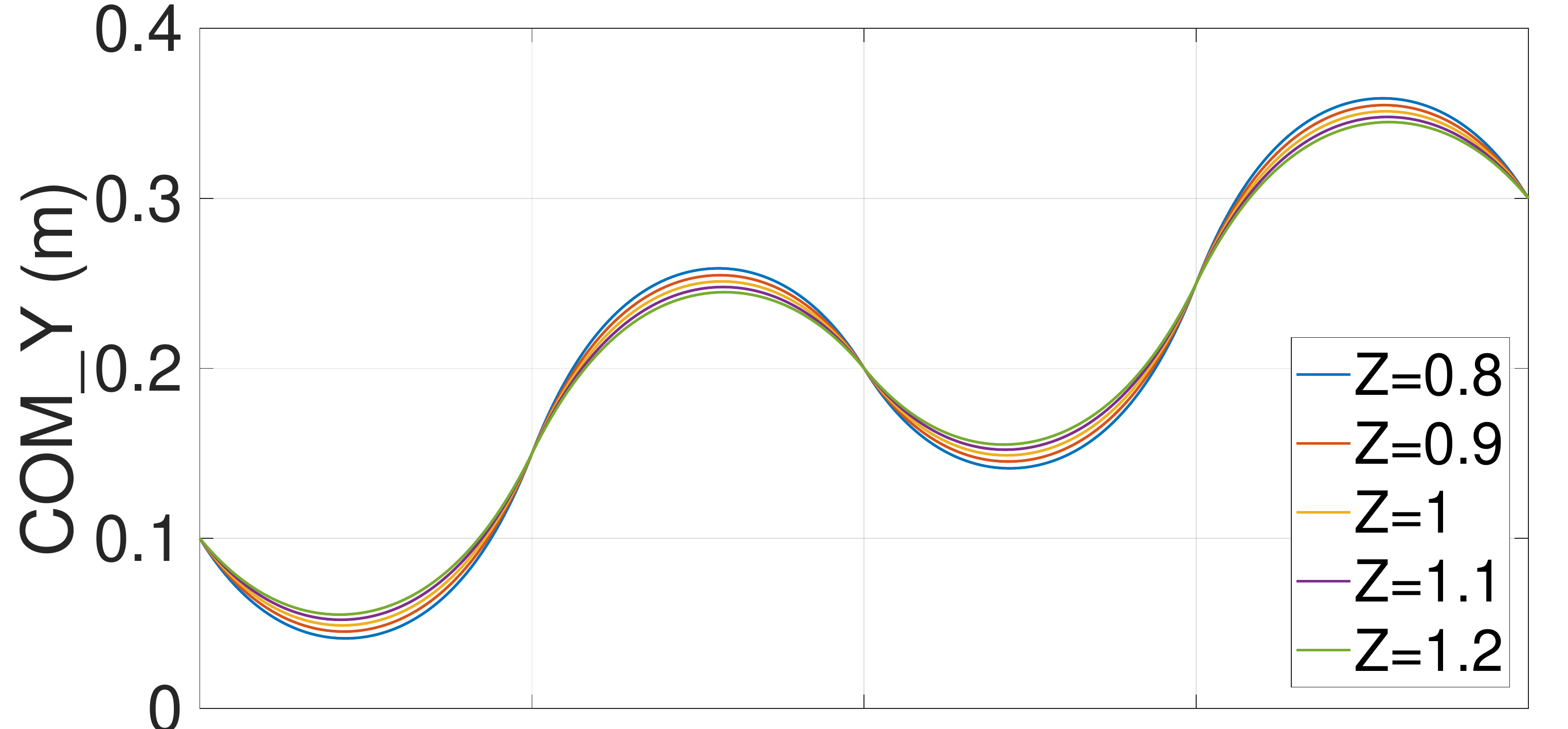}\\
			
			\includegraphics[width = 0.45\columnwidth, trim= 4.5cm 6cm 6.5cm 6.5cm,clip] {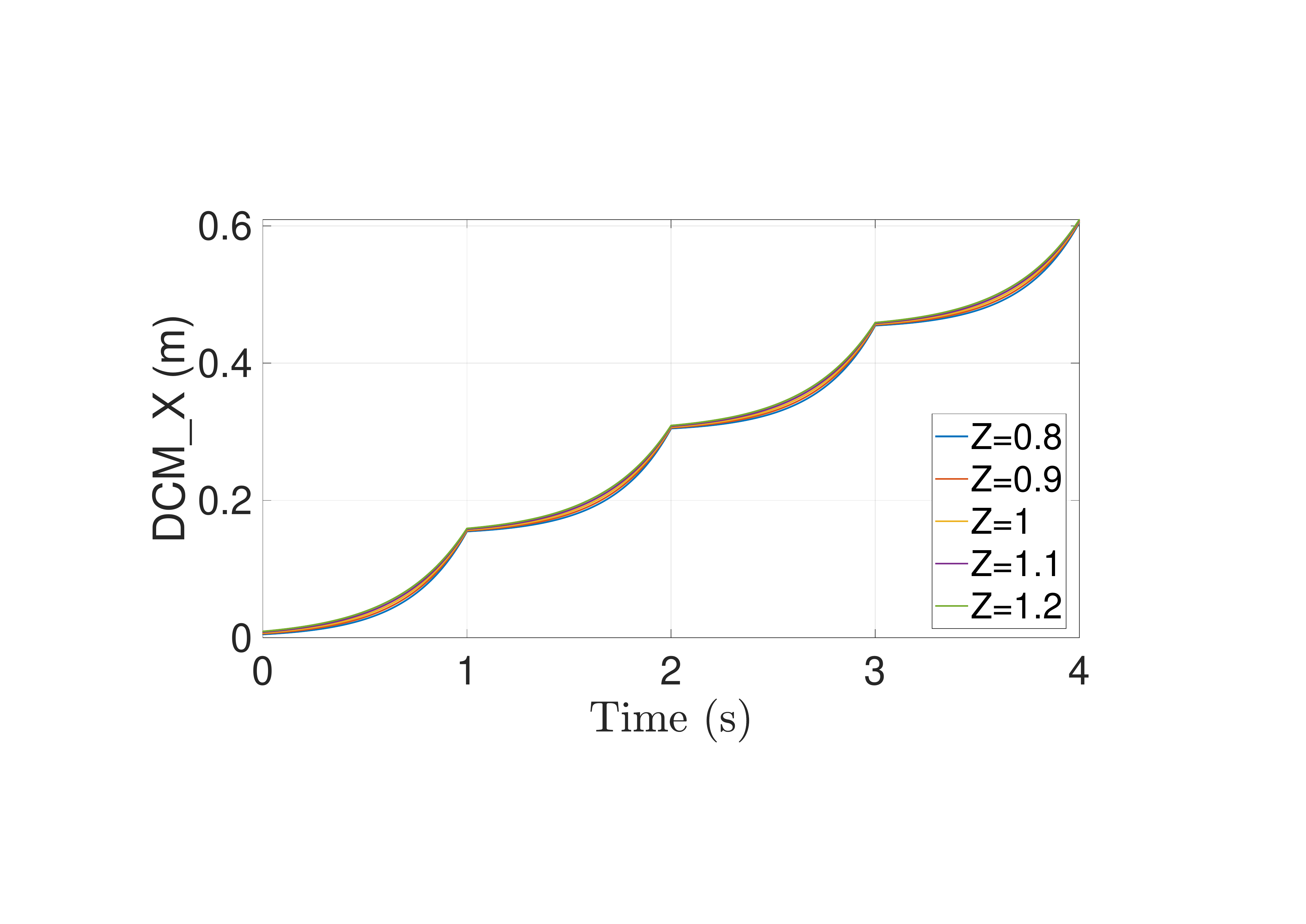}& 
			\includegraphics[width = 0.45\columnwidth, trim= 4.5cm 6cm 6.5cm 6.5cm,clip] {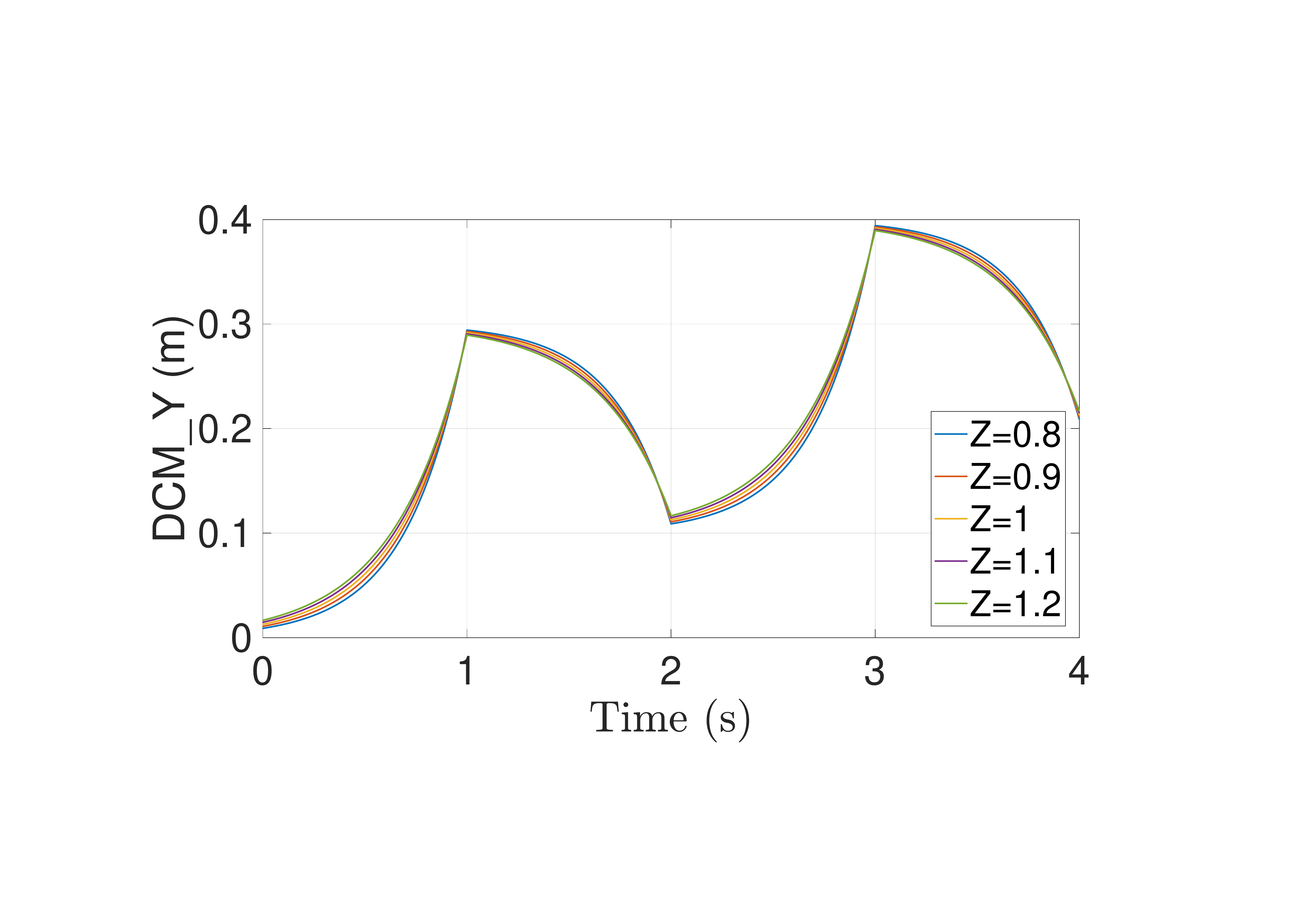}			
		\end{tabular}
	\end{centering}
	\caption{ A set of reference trajectories for a 4-steps diagonal walk with regarding to different height of COM ($z = 1\pm0.2 m$).}
	\label{fig:controller_diff_h}
\end{figure}
\begin{figure}[!t]
	\label {controller_sim_diff_h}
	\begin{centering}
		\begin{tabular}	{c c}			
			\includegraphics[width = 0.45\columnwidth, trim= 4.4cm 8.8cm 6.5cm 6.5cm,clip] {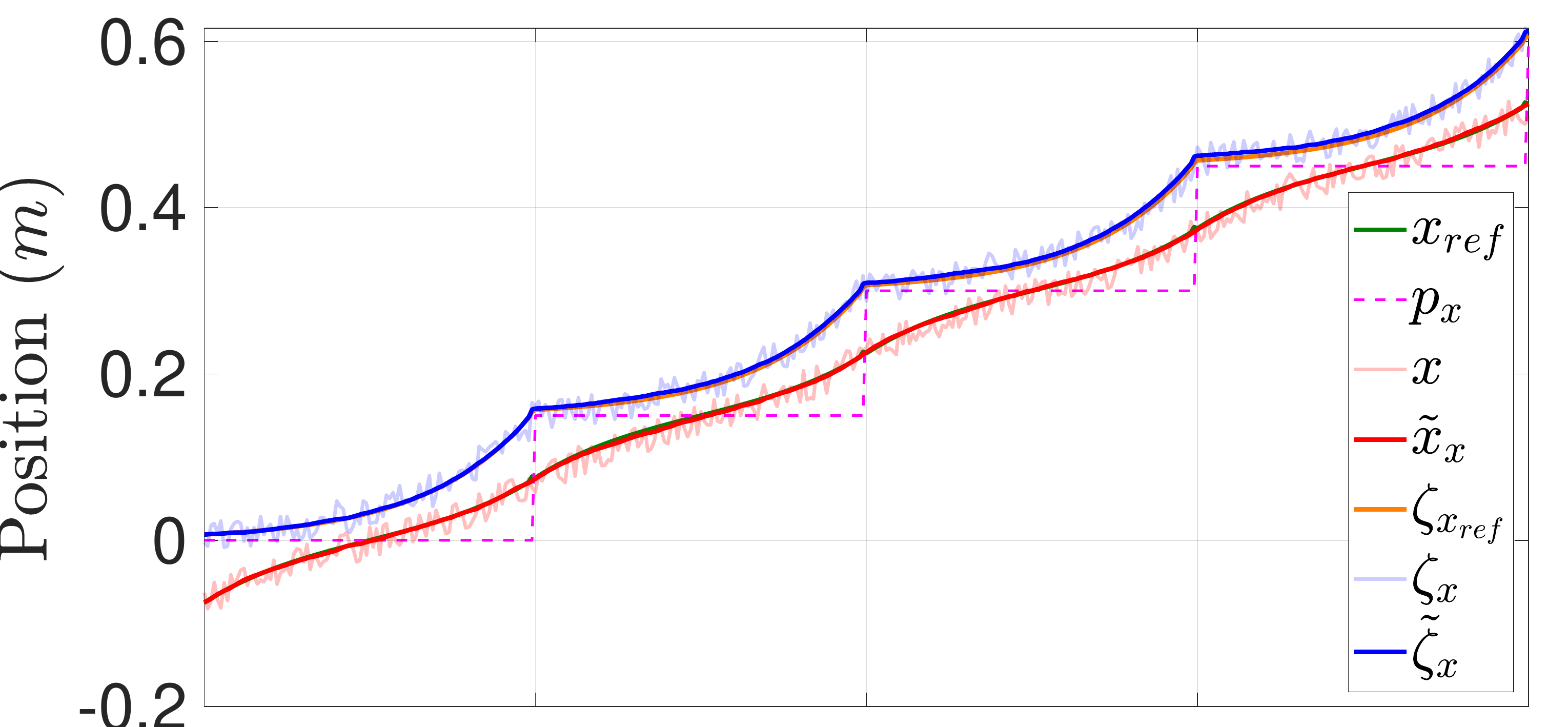}& 
			\includegraphics[width = 0.45\columnwidth, trim= 4.4cm 8.8cm 6.5cm 6.5cm,clip] {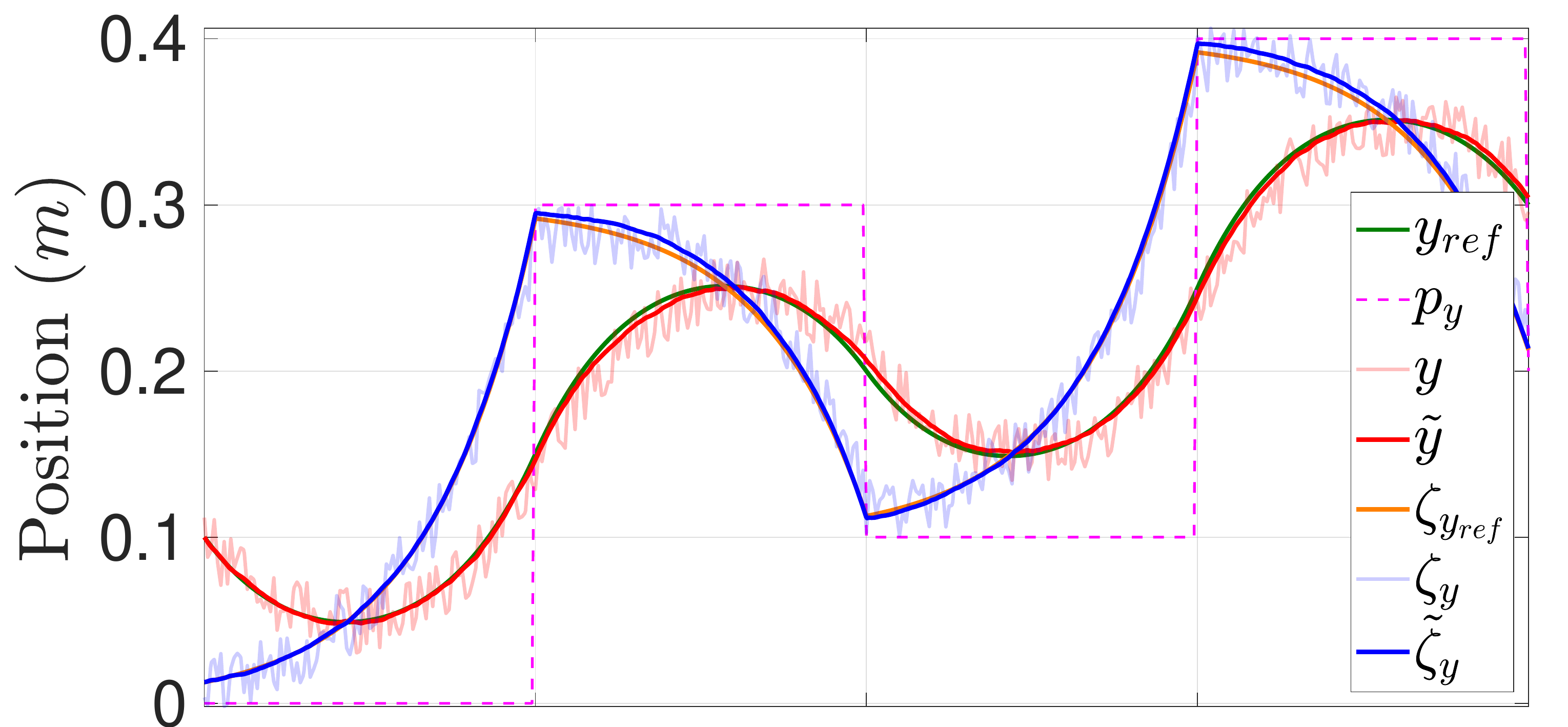}\\
			
			\includegraphics[width = 0.45\columnwidth, trim= 4.4cm 6cm 6.5cm 6.5cm,clip] {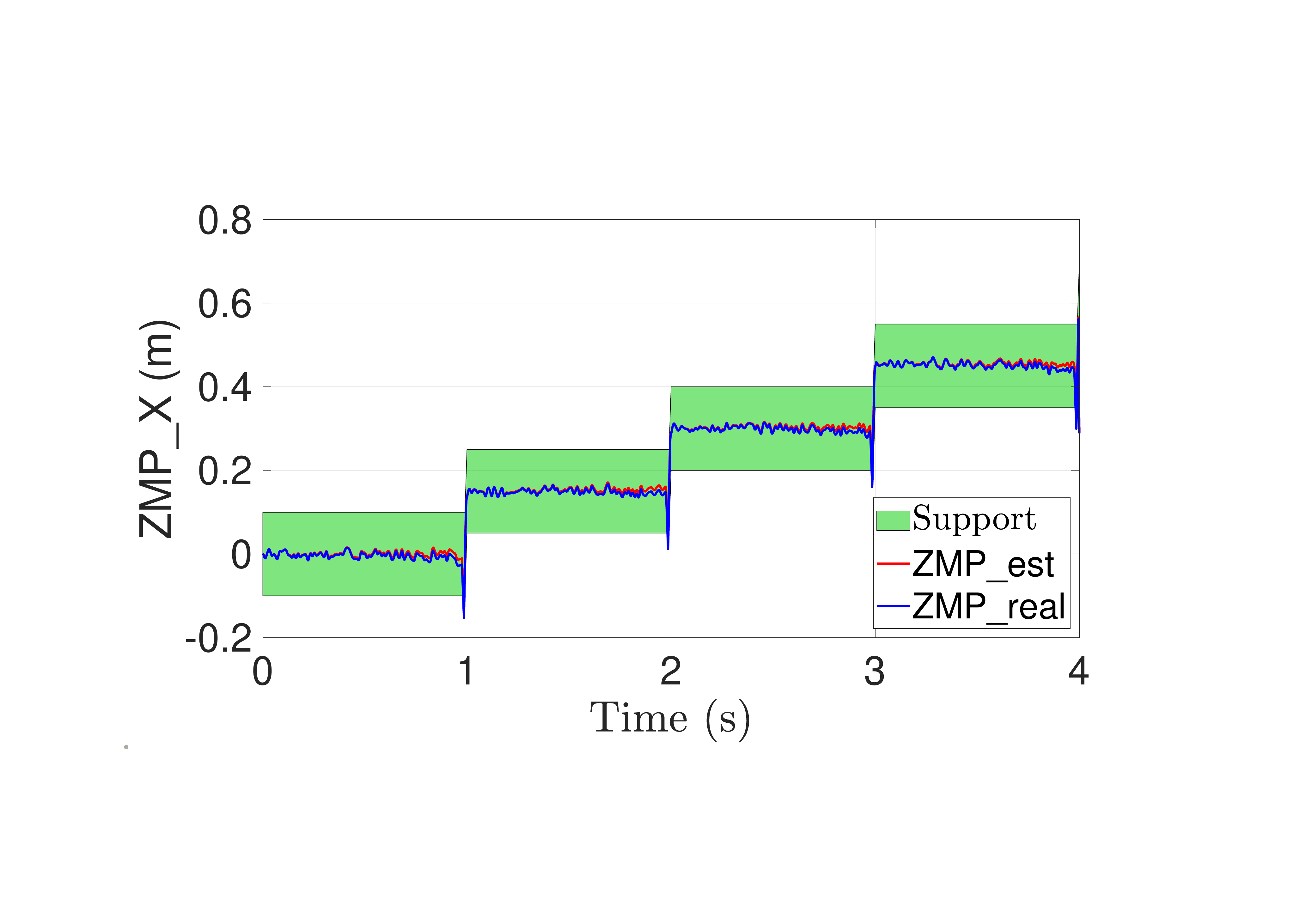}& 
			\includegraphics[width = 0.45\columnwidth, trim= 4.4cm 6cm 6.5cm 6.5cm,clip] {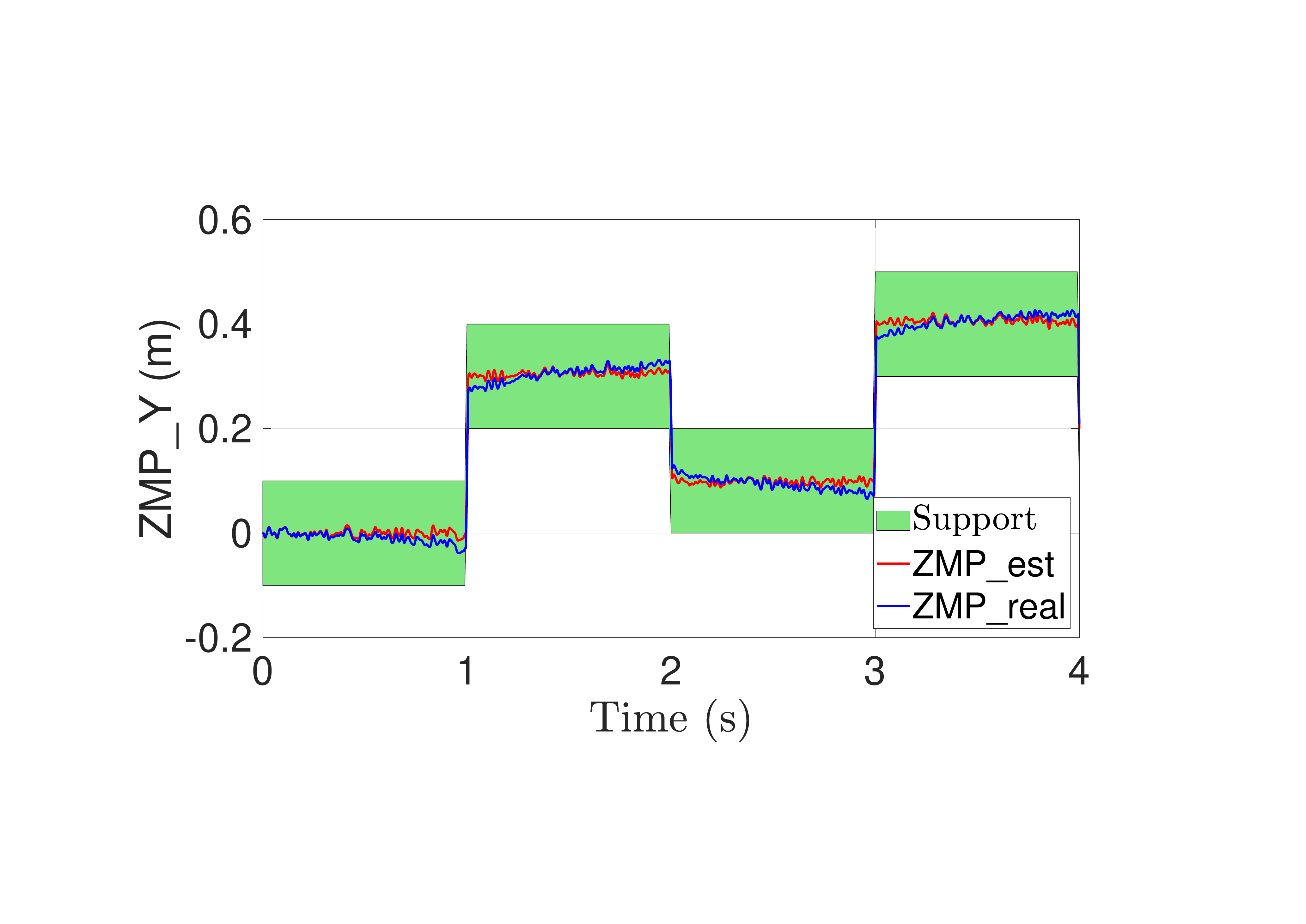}		
		\end{tabular}
	\end{centering}
	\caption{ Simulation results of examining the robustness w.r.t. COM height error. }
	\vspace{-5mm}
	\label{fig:controller_sim_diff_h}
\end{figure}

\textbf{Robustness w.r.t. COM height error:} Imagine a situation where a humanoid robot is carrying a heavy unknown weight object. In such situations, the height of the COM is changed, therefore the generated reference trajectories which are based on incorrect COM's height, may not guarantee the stability of the robot. To check this effect, a set of reference trajectories with regarding different height of COM ($z = z \pm 0.2 m$) has been generated and depicted in Fig.~\ref{fig:controller_diff_h}. As are shown in these plots, the incorrect height of COM has a significant effect in the frontal plane, but its effect in the sagittal plane is not too much. To examine the performance of the controller with regard to this type of error, a simulation has been set up. In this simulation, a simulated robot should walk diagonally and the actual height of COM is considered to be $z = 1.2m$ but the controller, state estimator and the generated reference trajectories are designed based on $z = 1m$. The estimated and real position of ZMP, COM, DCM have been recorded during simulation and shown in Fig.~\ref{fig:controller_sim_diff_h}. The results showed that although the controller tracks the inaccurate reference trajectories, ZMP still remains inside the support polygon and robot could perform stable walking.
\begin{figure}[!t]
	\centering
	\includegraphics[width=0.96\columnwidth, trim= 2cm 4.8cm 5cm 0.5cm,clip]{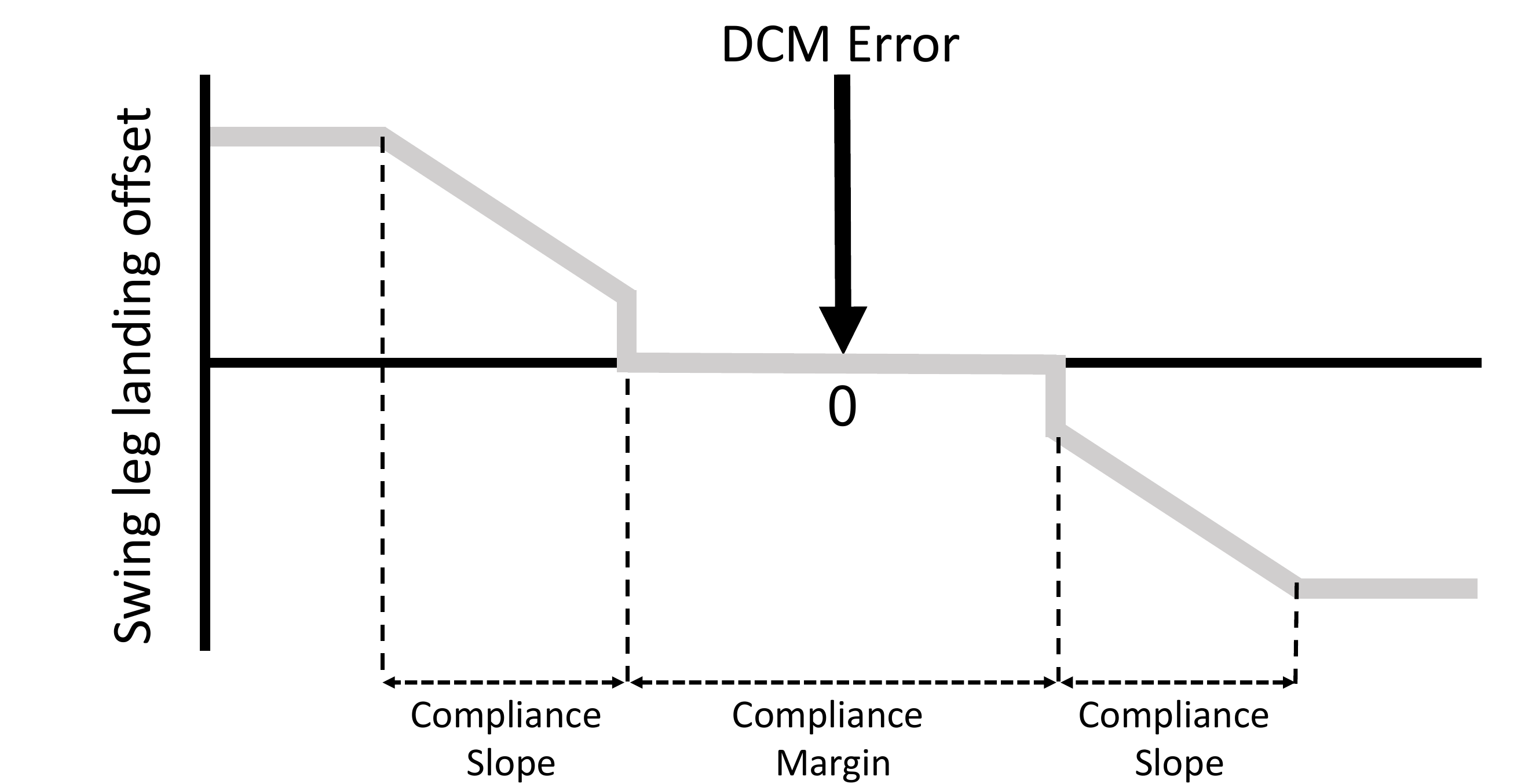}
	\caption{Adjusting the swing leg landing location based on DCM error.}		
	\vspace{-3mm}
	\label{fig:compliance}
\end{figure}
\section{Online Foot Step Adjustment}
\label{sec:online_footstep}
\begin{figure}[!t]
		\begin{centering}
		\begin{tabular}	{c c}			
						\includegraphics[height=3.1cm,width=0.52\columnwidth, trim= 4cm 6cm 6.2cm 7cm,clip]{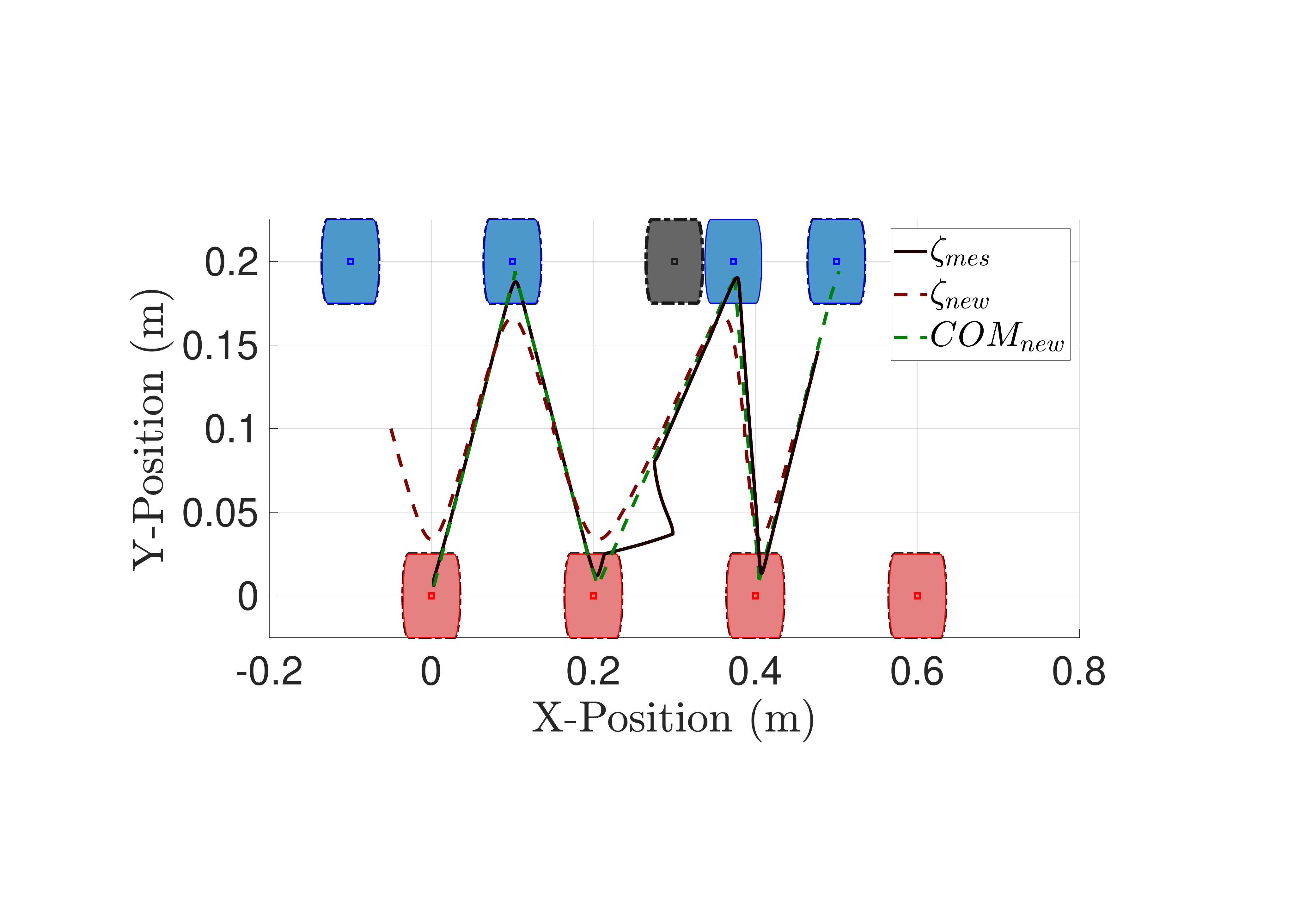} & \hspace{-3mm}
\includegraphics[scale=0.22, trim= 13cm 5cm 12cm 10cm,clip]{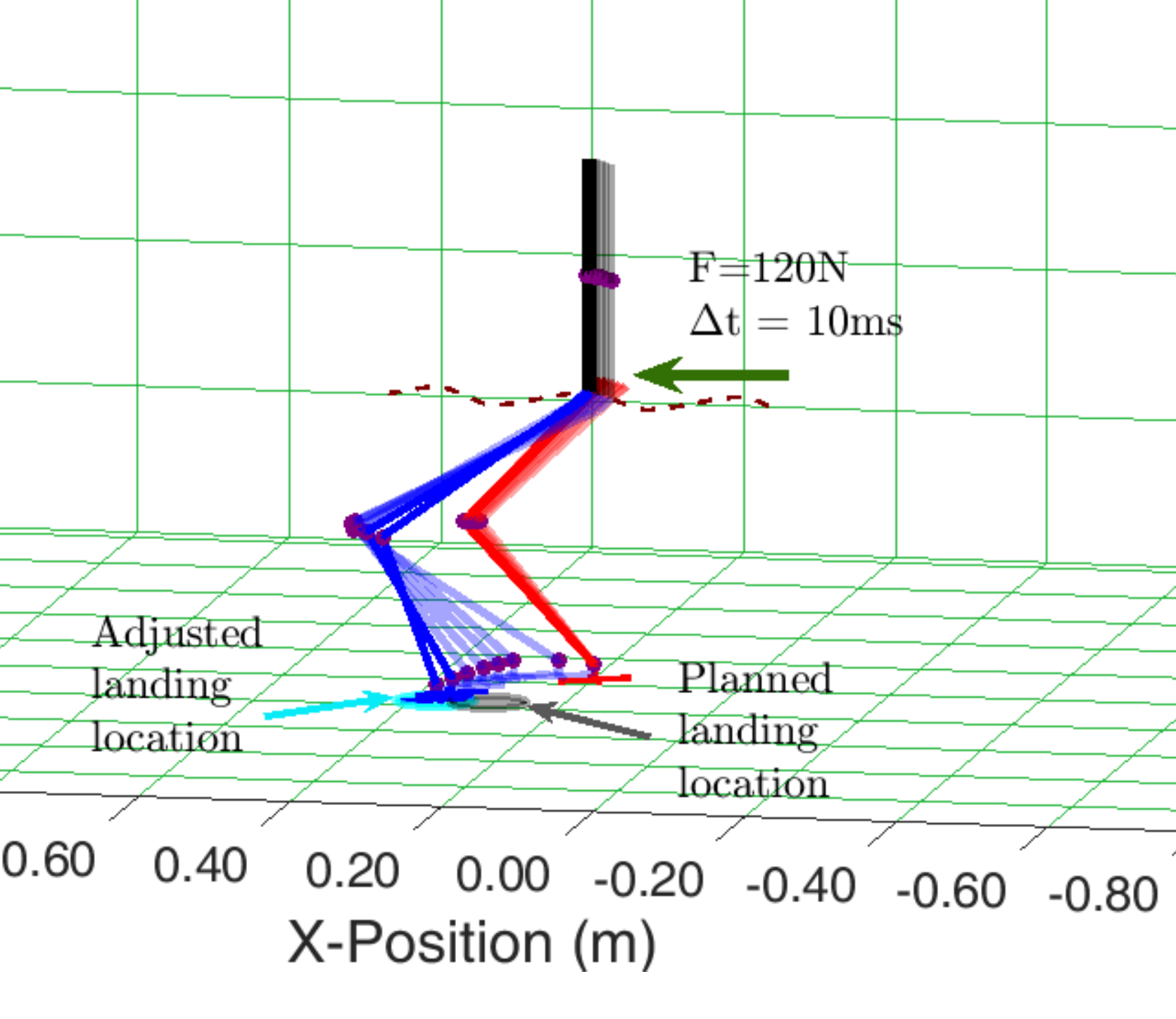}\\
(Top view) & (Side view)			
		\end{tabular}	
		\end{centering}
	\caption{ An example test of online foot step adjustment. The simulated robot is strongly pushed forward through the third step and the landing location of swing leg has been modified.}
	\vspace{-0mm}
	\label{fig:Push_DCM_exp}
\end{figure}

As shown in ZMP plots of Fig.~\ref{fig:controller_sim_ext_push_res}, in case of strong pushes, ZMP moves toward the edge of support polygon, as a consequence robot starts to roll over. In case of a more severe force ($F>90N$), the controller could not regain the stability of the robot because its output is bounded by the size of the support polygon. In such situations, the robot should change the landing location of the swing leg to regain its stability. According to the observability of DCM at each control cycle, the position of DCM at the end of the step can be predicted in advance by solving the DCM equation (Equation~\ref{eq:dcm}) as an initial value problem:
\begin{equation}
p_{x+1} = p_{x} + (\zeta_t - p_{x})e^{w(T-t)} \quad ,
\label{eq:dcm_at_end}
\end{equation}
\noindent
where $p_{x+1}$ is the next footstep position, $t, T, \zeta_t$ represent the time, step duration and measured DCM respectively. Based on this equation, the landing location of the swing leg can be adjusted by adding an offset according to the measured DCM at each control cycle. Unlike~\cite{diedam2008online,herdt2010online} which were based on optimization methods, using the difference between the desired and the predicted landing location, we define a compliance margin and a slope to calculate an offset for modifying the swing leg landing location. Fig.~\ref{fig:compliance} graphically shows how this offset is calculated based on the DCM error. Indeed, a compliance margin is considered to prevent unnecessary modification and a gain is used to define a compliance slope which is proportional to the DCM error. Moreover, the modification offset should be saturated in a certain value to keep the new landing position inside a kinematically reachable area of the robot. The compliance margin and slope are generally defined empirically. A simulation of online foot step adjustment is shown in Fig~\ref{fig:Push_DCM_exp}. In this simulation, the simulated robot should walk forward with a step length 0.2m and step duration 1s and while the robot is walking, at $t = 2.5s$, robot is subject to a severe forward push of $F=120N$ with impact duration of $\Delta t=10ms$, the compliance margin and slope are considered to be 0.025$m$ and $1.2$, respectively. As seen on Fig.~\ref{fig:Push_DCM_exp}, after applying this push, the DCM ($\zeta_{mes}$) is totally deviated from its desired trajectory which means the robot is starting to roll over and should change its swing leg landing position and re-plan all the reference trajectories to regain its stability. The simulation results showed that the controller could detect and react against this severe external push and could negate the effect of this disturbance. 
%
\section {Conclusion}
\label{sec:CONCLUSION}
In this paper, we have tackled the problem of designing an optimal closed-loop controller to provide a robust and efficient walking for biped robots. In particular, we used LIPM and DCM concepts to represent the overall dynamics model of a humanoid robot as a state space system and based on this system, we illustrated how the reference trajectories can be planned and controlled. Indeed, using this system, we formulated the problem of biped control as an LQG controller that provides optimal and robust solution using an offline optimization. Besides, we analyze the robustness of the proposed controller performing some simulations regarding three types of disturbances including measurement errors, unknown external disturbances, and COM height error. The simulation results showed that our controller is robust against these types of disturbances. Moreover, we proposed an online foot step adjustment method which was based on prediction of the DCM position at the end of each step. This method modifies the landing location of swing leg according to a compliance margin and a slope. Using a simulation scenario, we showed how this online modification could increase the withstanding level of the robot against external pushes.

In future work, we would like to port our proposed controller to real hardware to show the performance of the proposed system. Additionally, we would like to formulate our online foot step adjustment method as an online optimization problem. 

\section*{Acknowledgment}
This research is supported by Portuguese National Funds through Foundation for Science and Technology (FCT) through FCT scholarship SFRH/BD/118438/2016.

\bibliographystyle{IEEEtran}
\bibliography{IROS2019}


\end{document}